\documentclass{article}

\usepackage[final]{neurips_2021}

\usepackage[utf8]{inputenc} 
\usepackage[T1]{fontenc}    
\usepackage{hyperref}       
\usepackage{url}            
\usepackage{booktabs}       
\usepackage{amsfonts}       
\usepackage{nicefrac}       
\usepackage{microtype}      
\usepackage{xcolor}         

\usepackage{graphicx}
\usepackage{amsmath,amssymb,amsfonts,amsthm}
\usepackage{mathtools}
\usepackage{pifont}
\usepackage{nicefrac,xfrac}

\usepackage{wrapfig}
\usepackage{threeparttable}
\usepackage{cleveref}

\usepackage[para]{footmisc}

\makeatletter
\newcommand{\printfnsymbol}[1]{%
  \textsuperscript{\@fnsymbol{#1}}%
}
\makeatother

\hypersetup{
    colorlinks = true,
}

\newcommand{\x}{\mathbf{x}}

\usepackage{graphicx}
\usepackage{subcaption}
\usepackage{caption}

\newcommand{\Eb}[2]{\E_{#1}\!\left[#2\right]}

\newcommand{\bI}{\mathbf{I}}

\newcommand{\bx}{\mathbf{x}}

\newcommand{\bepsilon}{{\boldsymbol{\epsilon}}}
\newcommand{\bmu}{{\boldsymbol{\mu}}}

\newcommand\Tstrut{\rule{0pt}{2.6ex}}         

\newcommand\Tstrutsmall{\rule{0pt}{2.2ex}}         

\title{MCVD: Masked Conditional Video Diffusion for \\ Prediction, Generation, and Interpolation}

\author{
  Vikram Voleti\thanks{Equal Contribution} \\
  Mila, University of Montreal\\
  Canada \\
  \texttt{vikram.voleti@umontreal.ca} \\
   \And
   Alexia Jolicoeur-Martineau\printfnsymbol{1}  \\
   Mila, University of Montreal \\
   Canada \\
   \texttt{alexia.jolicoeur-martineau@mail.mcgill.ca} \\
   \AND
   Christopher Pal \\
   Mila, Polytechnique Montreal \\
   Canada CIFAR AI Chair\\
   ServiceNow Research \\
}

\usepackage{amsmath,amsfonts,bm}

\def\eqref#1{equation~\ref{#1}}

\def\1{\bm{1}}

\def\rvepsilon{{\bm{\epsilon}}}

\def\rvf{{\mathbf{f}}}

\def\rvp{{\mathbf{p}}}

\def\rvx{{\mathbf{x}}}

\def\rmI{{\mathbf{I}}}

\def\vzero{{\bm{0}}}

\DeclareMathAlphabet{\mathsfit}{\encodingdefault}{\sfdefault}{m}{sl}
\SetMathAlphabet{\mathsfit}{bold}{\encodingdefault}{\sfdefault}{bx}{n}

\def\gN{{\mathcal{N}}}

\newcommand{\E}{\mathbb{E}}

\usepackage{url}

\usepackage{color}
\usepackage{graphicx}
\usepackage{float}
\usepackage{subcaption}
\usepackage{multirow}
\usepackage{siunitx}
\newcommand{\Norm}[1]{\left\Vert#1\right\Vert}

\newcommand{\bq}{\begin{equation}}
\newcommand{\eq}{\end{equation}}

\newcommand{\balpha}{\bar\alpha}

\newcommand\numberthis{\addtocounter{equation}{1}\tag{\theequation}}

\allowdisplaybreaks

\begin{document}

\maketitle

\begin{abstract}

Video prediction is a challenging task. The quality of video frames from current state-of-the-art (SOTA) generative models tends to be poor and generalization beyond the training data is difficult. 
Furthermore, existing prediction frameworks are typically not capable of simultaneously handling other video-related tasks such as unconditional generation or interpolation. In this work, we devise a general-purpose framework called Masked Conditional Video Diffusion (MCVD) for all of these video synthesis tasks using a probabilistic conditional score-based denoising diffusion model, conditioned on past and/or future frames. We train the model in a manner where we randomly and independently mask all the past frames or all the future frames. This novel but straightforward setup allows us to train a single model that is capable of executing a broad range of video tasks, specifically: future/past prediction -- when only future/past frames are masked; unconditional generation -- when both past and future frames are masked; and interpolation -- when neither past nor future frames are masked. Our experiments show that this approach can generate high-quality frames for diverse types of videos. Our MCVD models are built from simple non-recurrent 2D-convolutional architectures, conditioning on blocks of frames and generating blocks of frames. We generate videos of arbitrary lengths autoregressively in a block-wise manner. Our approach yields SOTA results across standard video prediction and interpolation benchmarks, with computation times for training models measured in 1-12 days using $\le$ 4 GPUs. 

Project page: \url{https://mask-cond-video-diffusion.github.io}

Code: \url{https://mask-cond-video-diffusion.github.io/}
\end{abstract}

\section{Introduction}

\begin{figure}[h]
    \centering
    \includegraphics[width=.95\textwidth,interpolate=false]{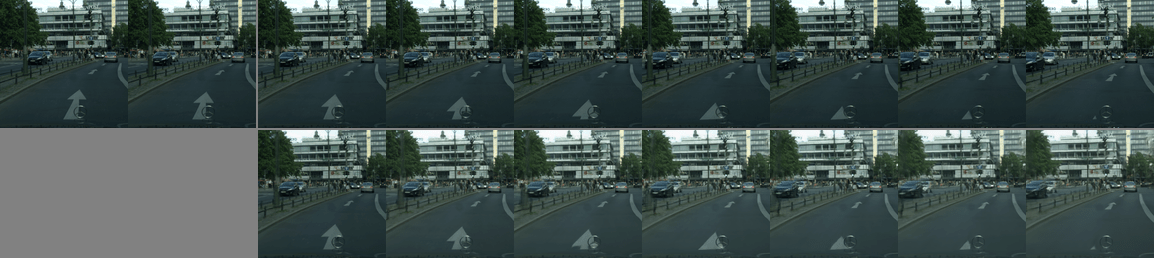}
    \includegraphics[width=.95\textwidth,interpolate=false]{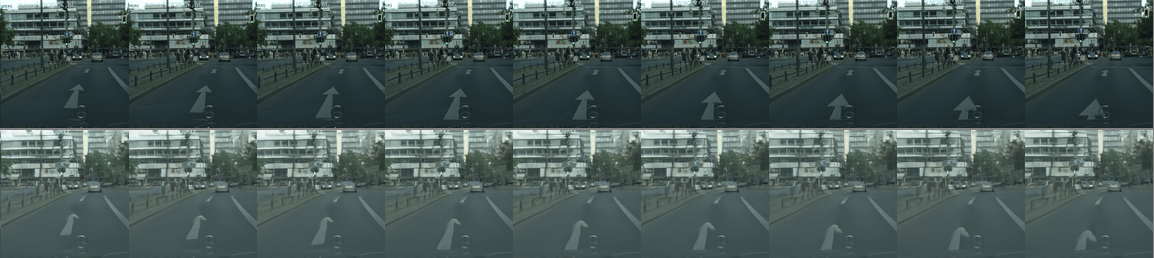}
    \caption{\small Our approach generates high quality frames many steps into the future: Given two conditioning frames from the Cityscapes \citep{cordts2016cityscapes} validation set (top left), we show 7 predicted future frames in row 2 below, then skip to frames 20-28, autoregressively predicted in row 4. Ground truth frames are shown in rows 1 and 3. Notice the initial large arrow advancing and passing under the car. In frame 20 (the far left of the 3rd and 4th row), the initially small and barely visible second arrow in the background of the conditioning frames has advanced into the foreground. Result generated by our \textbf{MCVD} concat model variant. Note that some Cityscapes videos contain brightness changes, which may explain the brightness change in this sample.   }
    \label{fig:teaser}
    \vspace{-0\baselineskip}
\end{figure}

Predicting what one may visually perceive in the future is closely linked to the dynamics of objects and people. As such, this kind of prediction relates to many crucial human decision-making tasks ranging from making dinner to driving a car. If video models could generate full-fledged videos in pixel-level detail with plausible futures, agents could use them to make better decisions, especially safety-critical ones. Consider, for example, the task of driving a car in a tight situation at high speed. Having an accurate model of the future could mean the difference between damaging a car or something worse. We can obtain some intuitions about this scenario by examining the predictions of our model in Figure \ref{fig:teaser}, where we condition on two frames and predict 28 frames into the future for a car driving around a corner. We can see that this is enough time for two different painted arrows to pass under the car. If one zooms in, one can inspect the relative positions of the arrow and the Mercedes hood ornament in the real versus predicted frames. Pixel-level models of trajectories, pedestrians, potholes, and debris on the road could one day improve the safety of vehicles.

Although beneficial to decision making, video generation is an incredibly challenging problem; not only must high-quality frames be generated, but the changes over time must be plausible and ideally drawn from an accurate and potentially complex distribution over probable futures. Looking far in time is exceptionally hard given the exponential increase in possible futures. Generating video from scratch or unconditionally further compounds the problem because even the structure of the first frame must be synthesized. 
Also related to video generation are the simpler tasks of a) video prediction, predicting the future given the past, and b) interpolation, predicting the in-between given past and future. Yet, both problems remain challenging. Specialized tools exist to solve the various video tasks, but they rarely solve more than one task at a time.



Given the monumental task of general video generation, current approaches are still very limited despite the fact that many state of the art methods have hundreds of millions of parameters \citep{wu2021greedy,weissenborn2019scaling,villegas2019high,babaeizadeh2021fitvid}. While industrial research is capable of looking at even larger models, current methods frequently underfit the data, leading to blurry videos, especially in the longer-term future and recent work has examined ways in improve parameter efficiency \citep{babaeizadeh2021fitvid}. Our objective here is to devise a video generation approach that generates high-quality, time-consistent videos within our computation budget of $\leq$ 4 GPU) and computation times for training models $\leq$ two weeks. Fortunately, diffusion models for image synthesis have demonstrated wide success, which strongly motivated our use of this approach. Our qualitative results in \autoref{fig:teaser} also indicate that our particular approach does quite well at synthesizing frames in the longer-term future (i.e., frame 29 in the bottom right corner).

One family of diffusion models might be characterized as Denoising Diffusion Probabilistic Models (DDPMs) \citep{sohl2015deep, ho2020denoising, dhariwal2021diffusion}, while another as Score-based Generative Models (SGMs) \citep{song2019generative, li2019learning, song2020improved, jolicoeur2020adversarial}. However, these approaches have effectively merged into a field we shall refer to as score-based diffusion models, which work by defining a stochastic process from data to noise and then reversing that process to go from noise to data. Their main benefits are that they generate very 1) high-quality and 2) diverse data samples. One of their drawbacks is that solving the reverse process is relatively slow, but there are ways to improve speed \citep{song2020ddim,jolicoeur2021gotta,salimans2022progressive,liu2022pseudo,xiao2021tackling}. Given their massive success and attractive properties, we focus here on developing our framework using score-based diffusion models for video prediction, generation, and interpolation.

Our work makes the following contributions:
\begin{enumerate}
    \item A conditional video diffusion approach for video prediction and interpolation that yields SOTA results.
    \item A conditioning procedure based on masking past and/or future frames in a blockwise manner giving a single model the ability to solve multiple video tasks: future/past prediction, unconditional generation, and interpolation. 
    \item A sliding window \emph{blockwise autoregressive} conditioning procedure to allow fast and coherent long-term generation (\autoref{fig:autoregression}).
    \item  A convolutional U-net neural architecture integrating recent developments with a conditional normalization technique we call SPAce-TIme-Adaptive Normalization (SPATIN) (\autoref{fig:our_net}).
\end{enumerate}

By conditioning on blocks of frames in the past and optionally blocks of frames even further in the future, we are able to better ensure that temporal dynamics are transferred across blocks of samples, i.e. our networks can learn \emph{implicit} models of spatio-temporal dynamics to inform frame generation. Unlike many other approaches, we do not have explicit model components for spatio-temporal derivatives or optical flow or recurrent blocks.  





\section{Conditional Diffusion for Video}

Let $\x_0 \in \mathbb{R}^d$ be a sample from the data distribution $p_{\text{data}}$. A sample $\x_0$ can corrupted from $t=0$ to $t=T$ through the Forward Diffusion Process (FDP) with the following transition kernel:
\begin{equation}
  q_t(\bx_t | \bx_{t-1}) = \mathcal{N}(\bx_t;\sqrt{1-\beta_t}\bx_{t-1},\beta_t \bI),
\end{equation}

Furthermore, $\x_t$ can be sampled directly from $\x_0$ using the following accumulated kernel:
\begin{align}
  q_t(\bx_t|\bx_0) = \mathcal{N}(\bx_t; \sqrt{\balpha_t}\bx_0, (1-\bar\alpha_t)\bI) \label{eq:q_marginal_arbitrary_t} \implies \rvx_t = \sqrt{\balpha_t}\rvx_0 + \sqrt{1 - \balpha_t} \rvepsilon
\end{align}
where $\bar\alpha_t = \prod_{s=1}^t (1 - \beta_s)$, and $\rvepsilon \sim \gN(\vzero, \rmI)$.

Generating new samples can be done by reversing the FDP  and solving the Reverse Diffusion Process (RDP) starting from Gaussian noise $\rvx_T$. It can be shown (\citet{song2020score, ho2020denoising}) that the RDP can be computed using the following transition kernel:
\begin{align*}
    &p_t(\bx_{t-1}|\bx_t,\bx_0) =  \mathcal{N}(\bx_{t-1}; \tilde\bmu_t(\bx_t, \bx_0), \tilde\beta_t \bI), \\
    \text{where}\quad \tilde\bmu_t(\bx_t, \bx_0) &= \frac{\sqrt{\bar\alpha_{t-1}}\beta_t }{1-\bar\alpha_t}\bx_0 + \frac{\sqrt{\alpha_t}(1- \bar\alpha_{t-1})}{1-\bar\alpha_t} \bx_t \quad \text{and} \quad
    \tilde\beta_t = \frac{1-\bar\alpha_{t-1}}{1-\bar\alpha_t}\beta_t
    \numberthis
    \label{eq:RDP}
\end{align*}

Since $\x_0$ given $\x_t$ is unknown, it can be estimated using eq. (\ref{eq:q_marginal_arbitrary_t}): $\hat\rvx_0 = \big(\rvx_t - \sqrt{1 - \balpha_t}\rvepsilon\big)/\sqrt{\balpha_t}$, where $\rvepsilon_\theta(\rvx_t|t)$ estimates $\rvepsilon$ using a time-conditional neural network parameterized by $\theta$. This allows us to reverse the process from noise to data. The loss function of the neural network is:
\begin{align}
 L(\theta) = \Eb{t, \bx_0 \sim p_{\text{data}}, \bepsilon \sim \mathcal{N}(\vzero, \bI)}{ \left\| \bepsilon - \bepsilon_\theta(\sqrt{\bar\alpha_t} \bx_0 + \sqrt{1-\bar\alpha_t}\bepsilon \mid t) \right\|_2^2} \label{eq:training_objective_simple}
\end{align}

Note that estimating $\epsilon$ is equivalent to estimating a scaled version of the score function (i.e., the gradient of the log density) of the noisy data:
\begin{align}
&\nabla_{\rvx_t} \log q_t (\rvx_t \mid \rvx_0) = -\frac{1}{1 - \balpha_t}(\rvx_t - \sqrt{\balpha_t} \rvx_0) = -\frac{1}{\sqrt{1 - \balpha_t}}\rvepsilon
\end{align}

Thus, data generation through denoising depends on the score-function, and can be seen as noise-conditional score-based generation.

Score-based diffusion models can be straightforwardly adapted to video by considering the joint distribution of multiple continuous frames. While this is sufficient for unconditional video generation, other tasks such as video interpolation and prediction remain unsolved. A conditional video prediction model can be approximately derived from the unconditional model using imputation~\citep{song2020score}; indeed, the contemporary work of \cite{ho2022VDM} attempts to use this technique; however, their approach is based on an approximate conditional model.

\subsection{Video Prediction via Conditional Diffusion}

We first propose to directly model the conditional distribution of video frames in the immediate future given past frames. Assume we have $p$ past frames $\rvp = \left\{ \rvp^i \right\}_{i=1}^{p}$ and $k$ current frames in the immediate future $\x_0 = \left\{ \x_0^i \right\}_{i=1}^{k}$. We condition the above diffusion models on the past frames to predict the current frames:
\begin{align}\label{eqn:vidpred}
 L_{\text{vidpred}}(\theta) = \Eb{t, [\rvp, \x_0] \sim p_{\text{data}}, \bepsilon \sim \mathcal{N}(\vzero, \bI)} {\Norm{ \bepsilon - \bepsilon_\theta(\sqrt{\balpha_t} \x_0 + \sqrt{1-\bar\alpha_t}\bepsilon \mid \rvp,  t)}^2}
\end{align}

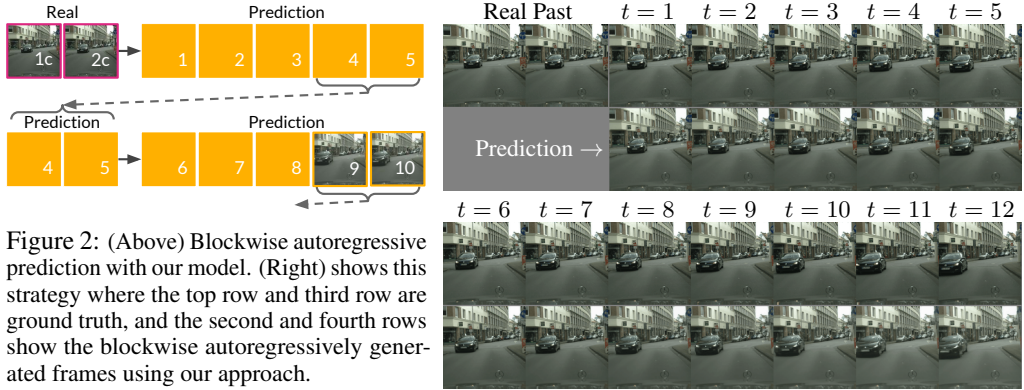
\begin{figure}[ht]
\begin{minipage}[b]{0.4\linewidth}
\centering
\includegraphics[width=\textwidth]{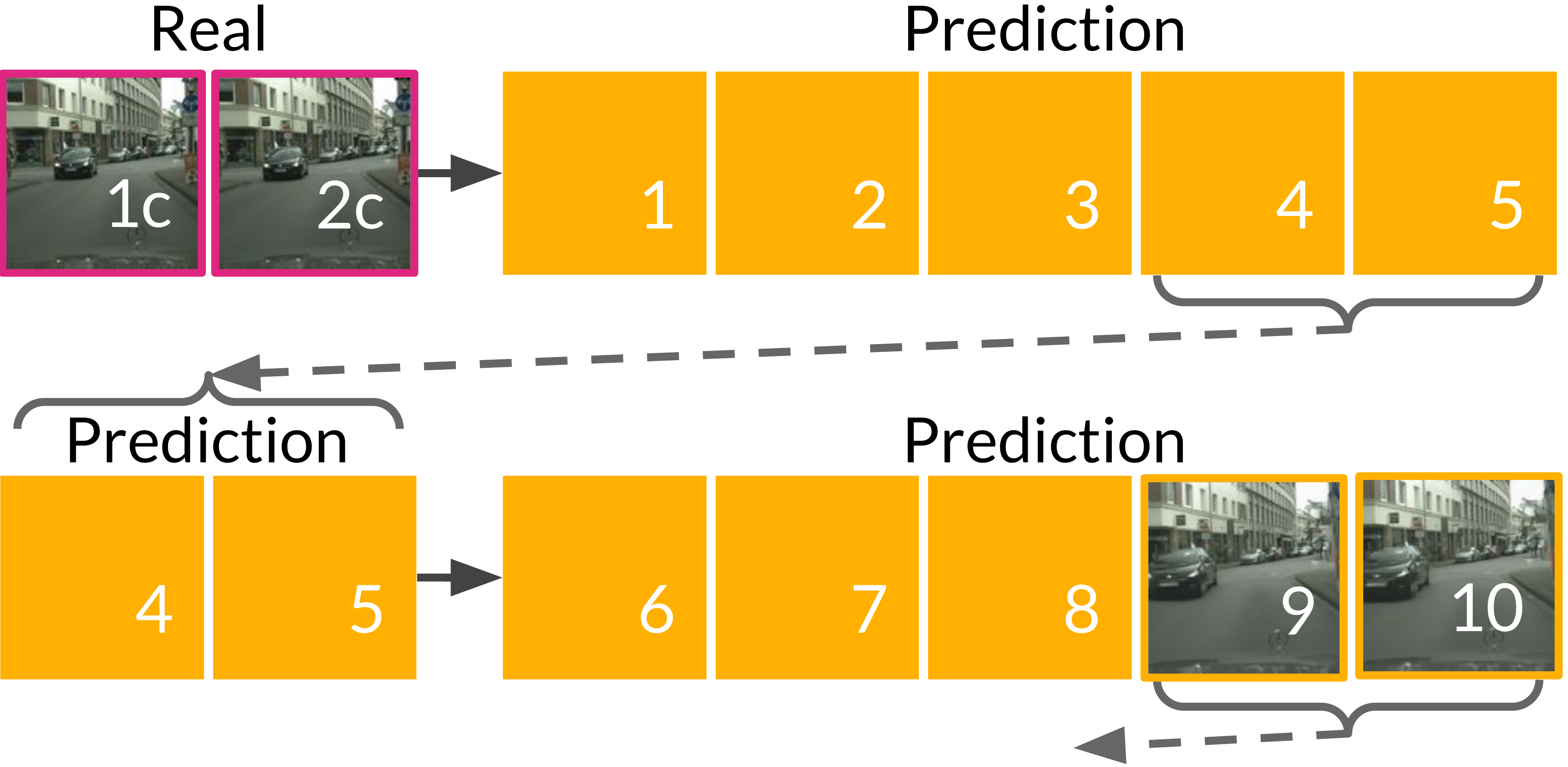}
    \caption{\small (Above) Blockwise autoregressive prediction with our model. 
    (Right) shows this strategy where the top row and third row are ground truth, and the second and fourth rows show the blockwise autoregressively generated frames using our approach. 
    }
\label{fig:autoregression}
\end{minipage}
\hspace{0.25cm}
\begin{minipage}[b]{0.55\linewidth}
\centering
\begin{picture}(150,150)
\put(-25, 140){\footnotesize Real Past}
\put(27,140){\footnotesize $t=1$}
\put(58,140){\footnotesize $t=2$}
\put(89,140){\footnotesize $t=3$}
\put(120,140){\footnotesize $t=4$}
\put(151,140){\footnotesize $t=5$}
\put(-40,75)
    {\includegraphics[width=1.0\textwidth,interpolate=false]{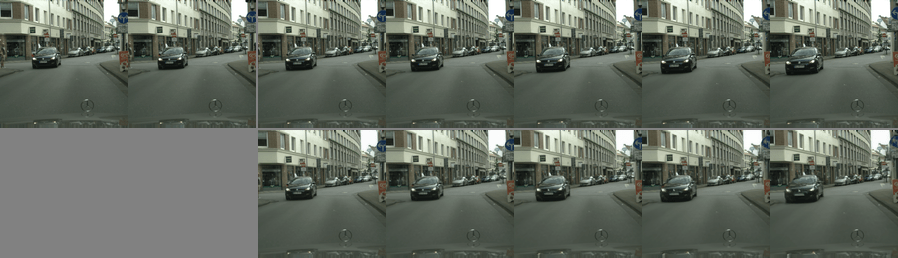}}
\put(-40,0)
    {\includegraphics[width=1.0\textwidth,interpolate=false]{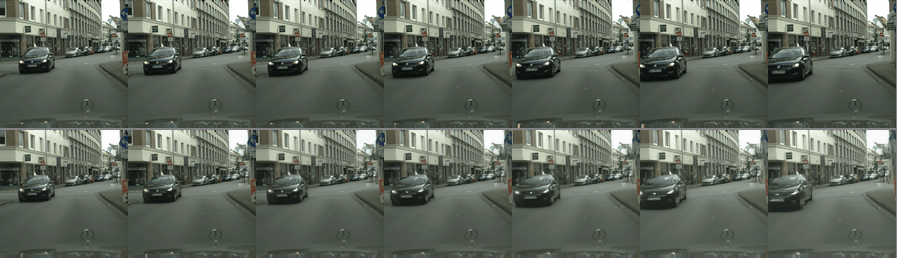}}
\put(-35, 65){\footnotesize $t=6$}
\put(-4, 65){\footnotesize $t=7$}
\put(27,65){\footnotesize $t=8$}
\put(58,65){\footnotesize $t=9$}
\put(89,65){\footnotesize $t=10$}
\put(120,65){\footnotesize $t=11$}
\put(151,65){\footnotesize $t=12$}
\put(-28, 88){\color{white}\footnotesize Prediction $\rightarrow$}
\end{picture}
\end{minipage}
\end{figure}

\begin{figure}
    \centering
    \includegraphics[width=0.9\textwidth]{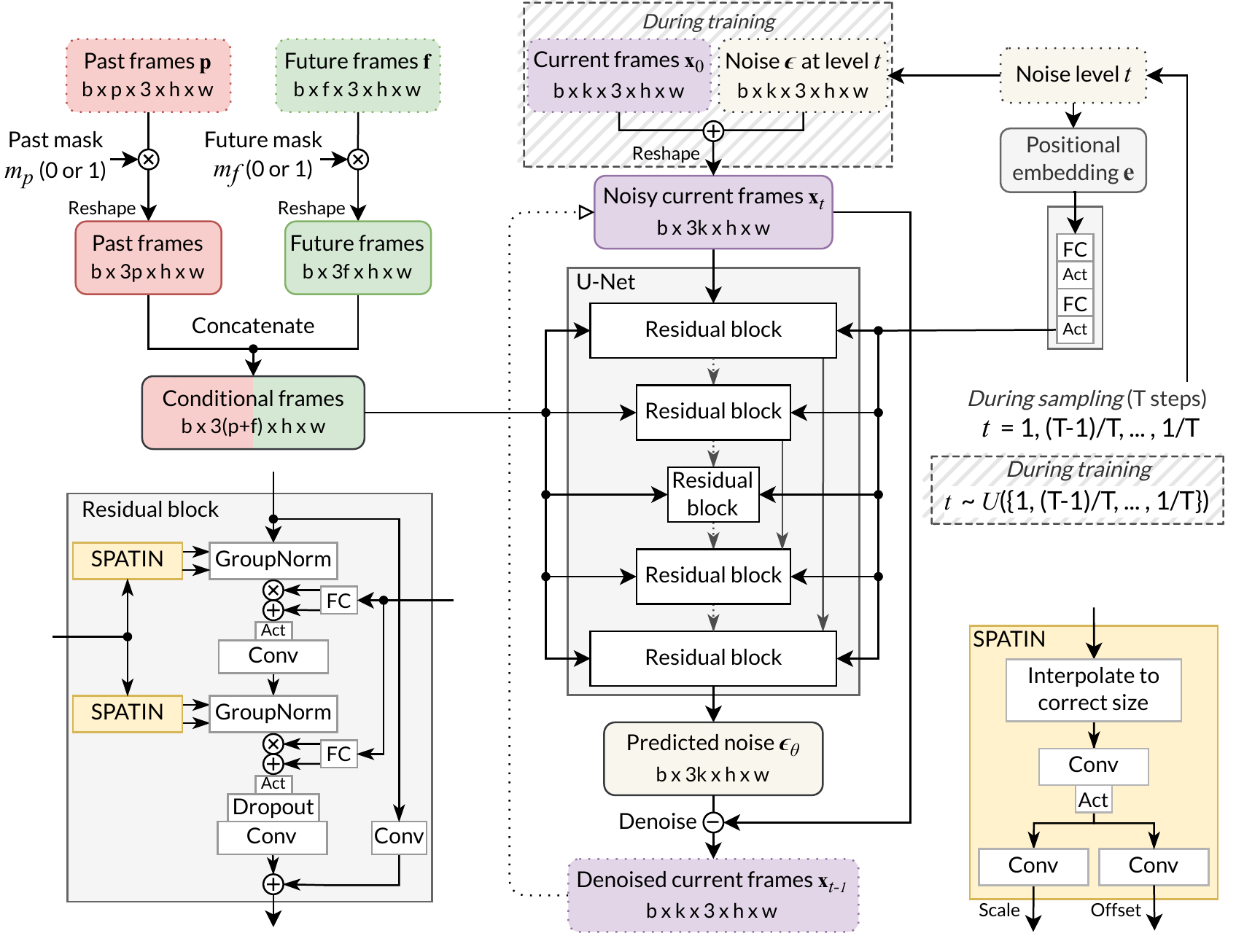}
    \caption{We give noisy current frames to a U-Net whose residual blocks receive conditional information from past/future frames and noise-level. The output is the predicted noise in the current frames, which we use to denoise the current frames. At test time, we start from pure noise.}
    \label{fig:our_net}
    \vspace{-1em}
\end{figure}

Given a model trained as above, video prediction for subsequent time steps can be achieved by blockwise autoregressively predicting current video frames conditioned on previously predicted frames (see \autoref{fig:autoregression}). We use variants of the network shown in \autoref{fig:our_net} to model $\bepsilon_\theta$ in \autoref{eqn:vidpred} here, and for \autoref{eqn:vidgen} and \autoref{eqn:vidgeneral} below.

\subsection{Video Prediction + Generation via Masked Conditional Diffusion}

Our approach above allows video prediction, but not unconditional video generation. As a second approach, we extend the same framework to video generation by masking (zeroing-out) the past frames with probability $p_{\text{mask}}=\nicefrac{1}{2}$ using binary mask $m_p$. The network thus learns to predict the noise added without any past frames for context. Doing so means that we can perform conditional as well as unconditional frame generation, i.e., video prediction and generation with the same network. This leads to the following loss ($\mathcal{B}$ is the Bernouilli distribution):
\begin{align}\label{eqn:vidgen}
 L_{\text{vidgen}}(\theta) = \Eb{t, [\rvp, \x_0] \sim p_{\text{data}}, \bepsilon \sim \mathcal{N}(\vzero, \bI), m_p \sim \mathcal{B}(p_{\text{mask}})}{ \left\| \bepsilon - \bepsilon_\theta(\sqrt{\bar\alpha_t} \x_0 + \sqrt{1-\bar\alpha_t}\bepsilon \mid m_p\rvp,  t) \right\|^2}
\end{align}
We hypothesize that this dropout-like \citep{srivastava14a} approach will also serve as a form of regularization, improving the model's ability to perform predictions conditioned on the past. We see positive evidence of this effect in our experiments -- see the MCVD past-mask model variants in \Cref{tab:bair_pred,tab:SMMNIST_ablation} versus without past-masking. Note that random masking is used only during training.
\subsection{Video Prediction + Generation + Interpolation via Masked Conditional Diffusion}

We now have a design for video prediction and generation, but it still cannot perform video interpolation nor past prediction from the future. As a third and final approach, we show how to build a general model for solving all four video tasks. Assume we have $p$ past frames, $k$ current frames, and $f$ future frames $\rvf = \left\{ \rvf^i \right\}_{i=1}^{f}$. We randomly mask the $p$ past frames with probability $p_{mask}=\nicefrac{1}{2}$, and similarly randomly mask the $f$ future frames with the same probability (but sampled separately). Thus, future or past prediction is when only future or past frames are masked. Unconditional generation is when both past and future frames are masked. Video interpolation is when neither past nor future frames are masked. The loss function for this general video machinery is:
\begin{align}\label{eqn:vidgeneral}
 L(\theta) = \Eb{t, [\rvp, \rvx_0, \rvf] \sim p_{\text{data}}, \bepsilon \sim \mathcal{N}(\vzero, \bI), ( m_p, m_f ) \sim \mathcal{B}(p_{\text{mask}})}{ \left\| \bepsilon - \bepsilon_\theta(\sqrt{\bar\alpha_t} \x_0 + \sqrt{1-\bar\alpha_t}\bepsilon \mid m_p\rvp, m_f\rvf,  t) \right\|^2}
\end{align}

\subsection{Our Network Architecture}

For our denoising network we use a U-net architecture \citep{ronneberger2015u, honari2016recombinator, salimans2017pixelcnn++} combining the improvements from \citet{song2020score} and \citet{dhariwal2021diffusion}. This architecture uses a mix of 2D convolutions \citep{fukushima1982neocognitron}, multi-head self-attention \citep{cheng2016long}, and adaptive group-norm \citep{wu2018group}. We use positional encodings of the noise level ($t \in [0,1]$) and process it using a transformer style positional embedding:
%
\vspace{-.2cm}
\begin{equation}
\textbf{e}(t)=\left[ \ldots,\cos \left(t c^{\frac{-2d}{D}} \right),  \sin \left(t c^{\frac{-2d}{D}} \right)
, \ldots  \right]^{\mathrm{T}},
\end{equation}
where $d=1,\ldots,D/2$ , $D$ is the number of dimensions of the embedding, and $c=10000$. This embedding vector is passed through a fully connected layer, followed by an activation function and another fully connected layer. Each residual block has an fully connected layer that adapts the embedding to the correct dimensionality.

To provide $\x_t$, $\rvp$, and $\rvf$ to the network, we separately concatenate the past/future conditional frames and the noisy current frames in the channel dimension. 
The concatenated noisy current frames are directly passed as input to the network. Meanwhile, the concatenated conditional frames are passed through an embedding that influences the conditional normalization akin to SPatially-Adaptive (DE)normalization (SPADE) \citep{park2019semantic}; to account for the effect of time/motion, we call this approach SPAce-TIme-Adaptive Normalization (SPATIN).
%
%
%
%
%
%
%
In addition to SPATIN, we also try directly concatenating the conditional and noisy current frames together and passing them as the input. In our experiments below we show some results with SPATIN and some with concatenation (concat).
%
%
%
%
For simple video prediction with \autoref{eqn:vidpred}, we experimented with 3D convolutions and 3D attention 
However, this requires an exorbitant amount of memory, and we found no benefit in using 3D layers over 2D layers at the same memory (i.e., the biggest model that fits in 4 GPUs). Thus, we did not explore this idea further. We also tried and found no benefit from gamma noise \citep{nachmani2021denoising}, L1 loss, and F-PNDM sampling \citep{liu2022pseudo}.


\section{Related work}


Score-based diffusion models have been used for image editing~\citep{meng2022sdedit, saharia2021palette, nichol2021glide} and our approach to video generation might be viewed as an analogy to classical image inpainting, but in the temporal dimension. 
The GLIDE or Guided Language to Image Diffusion for Generation and Editing approach of ~\cite{nichol2021glide} uses CLIP-guided diffusion for image editing, while Denoising Diffusion Restoration Models (DDRM)~\cite{kawar2022Ddrm} additionally condition on a corrupted image to restore the clean image. Adversarial variants of score-based diffusion models have been used to enhance quality \citep{jolicoeur2020adversarial} or speed \citep{xiao2021tackling}.

Contemporary work to our own such as that of ~\cite{ho2022VDM} and \cite{yang2022ResidualVideoDiffusion} also examine video generation using score-based diffusion models. However, the Video Diffusion Models (VDMs) work of~\citet{ho2022VDM} approximates conditional distributions using a gradient method for conditional sampling from their unconditional model formulation. In contrast, our approach directly works with a conditional diffusion model, which we obtain through masked conditional training, thereby giving us the exact conditional distribution as well as the ability to generate unconditionally. Their experiments focus on: a) unconditional video generation, and b) text-conditioned video generation, whereas our work focuses primarily on predicting future video frames from the past, using our masked conditional generation framework.  The Residual Video Diffusion (RVD) of ~\cite{yang2022ResidualVideoDiffusion} is only for video prediction, and it uses a residual formulation to generate frames autoregressively one at a time. Meanwhile, ours directly models the conditional frames to generate multiple frames in a block-wise autoregressive manner.

\begin{wrapfigure}[21]{R}{0.5\textwidth}
\vspace{-.5cm}
    \includegraphics[width=0.5\textwidth]{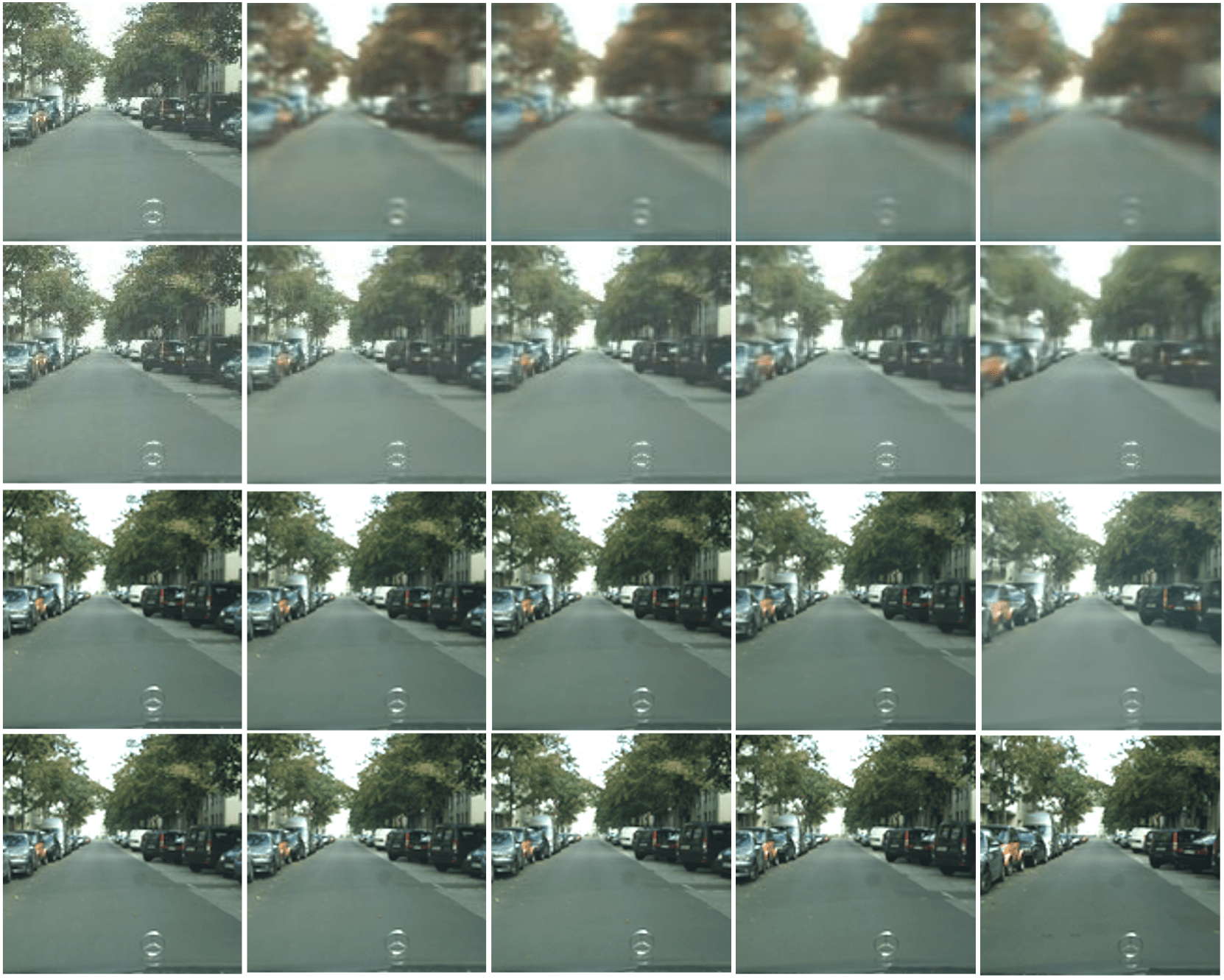}
    \caption{Comparing future prediction methods on Cityscapes:  SVG-LP (Top Row), Hier-vRNNs (Second Row), Our Method (Third Row), Ground Truth (Bottom Row). Frame 2, a ground truth conditioning frame is shown in first column, followed by frames: 3, 5, 10 and 20 generated by each method vs the ground truth at the bottom. 
    }
    \label{fig:RNN_compare}
\end{wrapfigure}

Recurrent neural network (RNN) techniques were early candidates for modern deep neural architectures for video prediction and generation. Early work combined RNNs with a stochastic latent variable (SV2P) \cite{babaeizadeh2018stochastic} and was optimized by variational inference. 
The stochastic video generation (SVG) approach of \cite{denton2018SVGLP} learned both prior and a per time step latent variable model, which influences the dynamics of an LSTM at each step. The model is also trained in a manner similar to a variational autoencoder, i.e., it was another form of variational RNN (vRNN). To address the fact that vRNNs tend to lead to blurry results, \cite{Castrejn2019ImprovedCV} (Hier-vRNN) increased the expressiveness of the latent distributions using a hierarchy of latent variables. We compare qualitative result of SVG and Hier-vRNN with the \textbf{MCVD} concat variant of our method in \Cref{fig:RNN_compare}. Other vRNN-based models include SAVP~\cite{Lee2018StochasticAV}, SRVP~\cite{franceschi2020stochastic}, SLAMP~\cite{akan2021slamp}. 

The well known Transformer paradigm \citep{vaswani2017attention} from natural language processing has also been explored for video. The Video-GPT work of \cite{yan2021videogpt} applied an autoregressive GPT style \citep{brown2020language} transformer to the codes produced from a VQ-VAE \citep{van2017neural}. 
The Video Transformer work of \cite{weissenborn2019scaling} models video using 3-D spatio-temporal volumes without linearizing positions in the volume. They examine local self-attention over small non-overlapping sub-volumes or 3D blocks. This is done partly to accelerate computations on TPU hardware.
%
%
Their work also observed that the peak signal-to-noise ratio (PSNR) metric and the mean-structural similarity (SSIM) metrics \citep{wang2004image} were developed for images, and have serious flaws when applied to videos. PSNR prefers blurry videos and SSIM does not correlate well to perceptual quality. Like them, we focus on the recently proposed Frechet Video Distance (FVD) \citep{unterthiner2018towards}, computed over entire videos and which is sensitive to visual quality, temporal coherence, and diversity of samples. \citet{rakhimov2020latent} (LVT) used transformers to predict the dynamics of video in latent space. \citet{le2021ccvs} (CCVS) also predict in latent space, that of an adversarially trained autoencoder, and also add a learnable optical flow module.

Generative Adversarial Network (GAN) based approaches to video generation have also been studied extensively. \citet{vondrick2016generating} proposed an early GAN architecture for video, using a spatio-temporal CNN.
\citet{villegas2017decomposing} proposed a strategy for separating motion and content into different pathways of a convolutional LSTM based encoder-decoder RNN.
\citet{saito2017temporal} (TGAN) predicted a sequence of latents using a temporal generator, and then the sequence of frames from those latents using an image generator. TGANv2~\cite{saito2020train} improved its memory efficiency.
MoCoGAN~\cite{tulyakov2018mocogan} explored style and content separation, but within a CNN framework. \citet{yushchenko2019markov} used the MoCoGAN framework by re-formulating the video prediction problem as a Markov Decision Process (MDP).
FutureGAN~\cite{aigner2018futuregan} used spatio-temporal 3D convolutions in an encoder decoder architecture, and elements of the progressive GAN~\cite{karras2018progressive} approach to improve image quality. TS-GAN~\cite{munoz2021temporal} facilitated information flow between consecutive frames. TriVD-GAN~\cite{luc2020transformation} proposes a novel recurrent unit in the generator to handle more complex dynamics, while DIGAN~\cite{yu2022generating} uses implicit neural representations in the generator.

Video interpolation was the subject of a flurry of interest in the deep learning community a number of years ago \citep{niklaus2017video, jiang2018super, xue2019video, bao2019depth}. However, these architectures tend to be fairly specialized to the interpolation task, involving optical flow or motion field modelling and computations. Frame interpolation is useful for video compression; therefore, many other lines of work have examined interpolation from a compression perspective. However, these architectures tend to be extremely specialized to the video compression task \citep{yang2020learning}.  

The Cutout approach of \cite{devries2017improved} has examined the idea of cutting out small continuous regions of an input image, such as small squares.
Dropout \citep{srivastava14a} at the FeatureMap level was proposed and explored under the name of SpatialDropout in \cite{tompson2015efficient}.
Input Dropout \citep{de2020input} has been examined in the context of dropping different channels of multi-modal input imagery, such as the dropping of the RGB channels or depth map channels during training, then using the model without one of the modalities during testing, e.g. in their work they drop the depth channel.

Regarding our block-autoregressive approach, previous video prediction models were typically either 1) non-recurrent: predicting all $n$ frames simultaneously with no way of adding more frames (most GAN-based methods), or 2) recurrent in nature, predicting 1 frame at a time in an autoregressive fashion. The benefit of the non-recurrent type is that you can generate videos faster than 1 frame at a time while allowing for generating as many frames as needed. The disadvantage is that it is slower than generating all frames at once, and takes up more memory and compute at each iteration. Our model finds a sweet spot in between in that it is block-autoregressive: generating $k < n$  frames at a time recurrently to finally obtain $n$ frames.

%
%
%

%
\section{Experiments}

We show the results of our \textbf{video prediction} experiments on test data that was never seen during training in Tables \ref{tab:SMMNIST}~-~\ref{tab:cityscape}
for Stochastic Moving MNIST (SMMNIST)~\footnote{\citep{denton2018SVGLP,srivastava2015unsupervised}},  KTH~\footnote{\citep{schuldt2004recognizing}},  BAIR~\footnote{\citep{ebert2017self}}, and Cityscapes~\footnote{\citep{cordts2016cityscapes}}respectively. We present \textbf{unconditional generation results} for BAIR in \autoref{tab:bair_gen} and UCF-101~\footnote{\citep{soomro2012ucf101}} in \autoref{tab:ucf_gen}, and \textbf{interpolation} results for SMMNIST, KTH, and BAIR in \autoref{tab:interp}.
 
\textbf{Datasets}: We generate 128x128 images for Cityscapes and 64x64 images for the other datasets. See our Appendix and supplementary material for additional visual results.  Our choice of datasets is in order of progressive difficulty: 1) SMMNIST: black-and-white digits; 2) KTH: grayscale single-humans; 3) BAIR: color, multiple objects, simple scene; 4) Cityscapes: color, natural complex natural driving scene; 5) UCF101: color, 101 categories of natural scenes. We process these datasets similarly to prior works. For Cityscapes, each video is center-cropped, then resized to $128\times128$. For UCF101, each video clip is center-cropped at 240×240 and resized to 64×64, taking care to maintain the train-test splits.

\begin{wraptable}[11]{htp}{.6\textwidth}
\small
\caption{Video prediction results on SMMNIST ($64\times64$) for 10 predicted frames conditioned on 5 past frames. We predicted 10 trajectories per real video, and report the average FVD and maximum SSIM, averaged across 256 test videos. }
\vspace{-0.8em}
\label{tab:SMMNIST}
\begin{tabular}{l|r|ll}
\toprule
\textbf{SMMNIST} [5 $\rightarrow$ 10; trained on $k$] & $k$ & \textbf{FVD$\downarrow$}      & \textbf{SSIM$\uparrow$}       \\ \hline
SVG \citep{denton2018SVGLP}
& 10 & 90.81                & 0.688     \Tstrut           \\
vRNN 1L \citep{Castrejn2019ImprovedCV}
& 10 & 63.81                & 0.763                \\
Hier-vRNN  \citep{Castrejn2019ImprovedCV}
& 10 & 57.17                & 0.760                \\
\textbf{MCVD} concat (Ours) & \textbf{5} & 25.63 & \textbf{0.786}    \\
\textbf{MCVD} spatin (Ours) & \textbf{5} & \textbf{23.86} & 0.780    \\
\bottomrule
\end{tabular}
\end{wraptable}

Unless otherwise specified, we set the mask probability to 0.5 when masking was used. For sampling, we report results using the sampling methods DDPM~\citep{ho2020denoising} or DDIM~\citep{song2020ddim} with only 100 sampling steps, though our models were trained with 1000, to make sampling faster. We observe that the metrics are generally better using DDPM than DDIM (except for UCF-101). Using 1000 sampling steps could yield better results.

Note that all our models are trained to predict only 4-5 current frames at a time, unlike other models that predict $\ge$10. We use these models to then autoregressively predict longer sequences for prediction or generation. This was done in order to fit the models in our GPU memory budget. Despite this disadvantage, we find that our MCVD models perform better than many previous SOTA methods. 
\vspace{1em}
\begin{wraptable}[13]{h}{9cm}
\vspace{-.45cm}
\centering
\small
\caption{Video prediction results on KTH ($64\times64$), predicting 30 and 40 frames using models trained to predict $k$ frames at a time. All models condition on 10 past frames, on 256 test videos.}
\vspace{-.5em}
\setlength{\tabcolsep}{2pt}
\begin{tabular}{l|rc|lcl}
\toprule
\textbf{KTH} [10 $\rightarrow$ $pred$; trained on $k$] & $k$ & $pred$ & FVD$\downarrow$    & PSNR$\uparrow$  & SSIM$\uparrow$  \\ \hline
SAVP~\citep{Lee2018StochasticAV}         & 10 & 30 & 374 $\pm$ 3 & 26.5  & 0.756 \Tstrutsmall \\
\textbf{MCVD} concat (Ours) & \textbf{5} & 30 & 323 $\pm$ 3 & 27.5   &  0.835    \\
SLAMP~\citep{akan2021slamp}              & 10 & 30 & 228 $\pm$ 5 & 29.4 & 0.865 \\
SRVP~\citep{franceschi2020stochastic}    & 10 & 30 & \textbf{222} $\pm$ 3 & \textbf{29.7} & \textbf{0.870} \\
\hline
\textbf{MCVD} concat (Ours)                                    & \textbf{5} & 40   & 276.7  &     26.40  &  0.812   \Tstrutsmall  \\
SAVP-VAE~\citep{Lee2018StochasticAV}                    & 10 & 40 & 145.7 & 26.00 & 0.806 \\
Grid-keypoints~\citep{gao2021accurate}                  & 10 & 40   & \textbf{144.2} & \textbf{27.11} & \textbf{0.837} \\
\bottomrule
\end{tabular}
\label{tab:KTH}
\end{wraptable}
\textbf{Metrics}: As mentioned earlier, we primarily use the FVD metric for comparison across models as FVD measures both fidelity and diversity of the generated samples. Previous works compare Frechet Inception Distance (FID)~\citep{heusel2017gans} and Inception Score (IS)~\citep{salimans2016improved}, adapted to videos by replacing the Inception network with a 3D-convolutional network that takes video input. FVD is computed similarly to FID, but using an I3D network trained on the huge video dataset Kinetics-400. We also report PSNR and SSIM.

\textbf{Ablation studies}: In \autoref{tab:bair_pred} we compare models that use concatenated raw pixels as input to U-Net blocks (concat) to SPATIN variants. We also compare no-masking to past-masking variants, i.e. models which are only trained predict the future vs. models which are regularized by being trained for prediction and unconditional generation. It can be seen that our model works across different choices of past frames and generates better quality for shorter videos. This is expected from models of this kind. Moreover, it can be seen that the model trained on the two tasks of Prediction and Generation (i.e., the models with past-mask) performs better than the model trained only on Prediction!

In addition, the appendix contains an ablation study in \autoref{tab:SMMNIST_ablation} on the different design choices: concat vs concat past-future-mask vs spatin vs spatin future-mask vs spatin past-future-mask. It can be seen that concat is, in general, better than spatin. It can also be seen that the past-future-mask variant, which is a general model capable of all three tasks, performs better at the individual tasks than the models trained only on the individual task. This was demonstrated in \autoref{tab:bair_pred} as well. This shows that the model gains very helpful insights while generalizing to all three tasks, which it does not while training only on the individual task.

We conducted preliminary experiments with a larger number of frames. Since the models with a larger number of frames were bigger, we could only run them for a shorter time with a smaller batch size than the smaller models. In general, we found that larger models did not substantially improve the results. We attribute this to the fact that using more frames means that the model should be given more capacity, but we could not increase it due to our computational budget constraints. We emphasize that our method works very well with fewer computational resources.


Examining these results we remark that we have SOTA performance for prediction on SMMNIST, BAIR and the challenging Cityscapes evaluation. Our Cityscapes model yields an FVD of 145.5, whereas the best previous result of which we are aware is 418. The quality of our Cityscapes results are illustrated visually in \autoref{fig:teaser} and \autoref{fig:autoregression} and in the additional examples provided 
in our Appendix. While our completely unconditional generation results are strong, we note that when past masking is used to regularize future predicting models, we see clear performance gains in \autoref{tab:bair_pred}. Finally, in \autoref{tab:interp} we see that our interpolation results are SOTA by a wide margin, across experiments on SMMNIST, KTH and BAIR -- even compared to architectures much more specialized for interpolation.

It can be seen that our proposed method generates better quality videos, even though it was trained on a shorter number of frames than other methods. It can also be seen that training on multiple tasks using random masking improves the quality of generated frames than training on the individual tasks.

\begin{threeparttable}[h]
    \small
	\caption{Video prediction results on  BAIR ($64\times64$) conditioning on $p$ past frames and predicting $pred$ frames in the future, using models trained to predict $k$ frames at at time. 
	}
	\label{tab:bair_pred}
	\centering
	\begin{tabular}{l|crc|rcc}
	\toprule
		\textbf{BAIR} ($64\times64$) [past $p$ $\rightarrow$ $pred$ ; trained on $k$] & $p$ & $k$ & $pred$ & FVD$\downarrow$ & PSNR$\uparrow$ & SSIM$\uparrow$ \\
		\cmidrule(){1-7}
        LVT \citep{rakhimov2020latent} & 1 & 15 & 15 & 125.8 & -- & -- \\
        DVD-GAN-FP \citep{clark2019adversarial} & 1 & 15 & 15 & 109.8 & -- & -- \\
		\textbf{MCVD} spatin (Ours) & 1 & \textbf{5} & 15 & 103.8 & 18.8 & 0.826  \\
        TrIVD-GAN-FP~\citep{luc2020transformation} & 1 & 15 & 15 & 103.3 & -- & -- \\
        VideoGPT~\citep{yan2021videogpt} & 1 & 15 & 15 & 103.3 & -- & -- \\
		CCVS \citep{le2021ccvs} & 1 & 15 & 15 & 99.0 & -- & -- \\
	    \textbf{MCVD} concat (Ours) & 1 & \textbf{5} & 15  & 98.8 & 18.8 & 0.829 \\
		\textbf{MCVD} spatin past-mask (Ours) & 1 & \textbf{5} & 15 & 96.5 & 18.8 & 0.828 \\
		\textbf{MCVD} concat past-mask (Ours) & 1 & \textbf{5} & 15 & 95.6 & 18.8 & \textbf{0.832}  \\
        Video Transformer \citep{weissenborn2019scaling} & 1 & 15 & 15 & 94-96\tnote{a} & -- & -- \\
        FitVid \citep{babaeizadeh2021fitvid} & 1 & 15 & 15 & 93.6 & -- & -- \\
		\textbf{MCVD} concat past-future-mask (Ours) & 1 & \textbf{5} & 15 & \textbf{89.5} & 16.9 & 0.780  \\
		\cmidrule(){1-7}
        SAVP \citep{Lee2018StochasticAV} & 2 & 14 & 14 & 116.4 & -- & -- \\
		\textbf{MCVD} spatin (Ours) & 2 & \textbf{5} & 14 & 94.1 & 19.1 & 0.836  \\
		\textbf{MCVD} spatin past-mask (Ours) & 2 & \textbf{5} & 14 & 90.5 & \textbf{19.2} & 0.837  \\
		\textbf{MCVD} concat (Ours)  & 2 & \textbf{5} & 14 & 90.5 & 19.1 & 0.834  \\
		\textbf{MCVD} concat past-future-mask (Ours) & 2 & \textbf{5} & 14 & 89.6 & 17.1 & 0.787  \\
		\textbf{MCVD} concat past-mask (Ours) & 2 & \textbf{5} & 14 & \textbf{87.9} & 19.1 & \textbf{0.838}  \\
		\cmidrule(){1-7}
        SAVP~\citep{Lee2018StochasticAV} & 2    & 10   & 28   & 143.4          &      --            & 0.795  \\
        Hier-vRNN ~\citep{Castrejn2019ImprovedCV}                             & 2    & 10   & 28   & 143.4           &           --      & \textbf{0.822}   \\
		
		\textbf{MCVD} spatin (Ours) & 2 & \textbf{5} & 28 & 132.1 & 17.5 & 0.779  \\
		\textbf{MCVD} spatin past-mask (Ours) & 2 & \textbf{5} & 28 & 127.9 & 17.7 & 0.789 \\
		\textbf{MCVD} concat (Ours)  & 2 & \textbf{5} & 28 & 120.6 & 17.6 & 0.785  \\
		\textbf{MCVD} concat past-mask (Ours) & 2 & \textbf{5} & 28 & 119.0 & \textbf{17.7} & 0.797 \\
		\textbf{MCVD} concat past-future-mask (Ours) & 2 & \textbf{5} & 28 & \textbf{118.4} & 16.2 & 0.745 \\
		\bottomrule
	\end{tabular}
    \begin{tablenotes}
    \item[a] 94 on only the first frames, 96 on all subsequences of test frames
    \end{tablenotes}
\end{threeparttable}

\begin{table}[h]
\centering
\small
\caption{Video prediction on Cityscapes ($128\times128$) conditioning on 2 frames and predicting 28. SPATIN seems to produce a drift towards brighter images with a color balance shift in frames further from the start frame on Cityscapes, resulting in increased FVD for SPATIN than the CONCAT variant.
}
\label{tab:cityscape}
\begin{tabular}{l|r|lll}
\toprule
\textbf{Cityscapes} ($128\times128$) [2 $\rightarrow$ 28; trained on $k$] & $k$ & FVD$\downarrow$      & LPIPS$\downarrow$ & SSIM$\uparrow$   \\ \hline
SVG-LP~\cite{denton2018SVGLP}     & 10 & 1300.26              & 0.549 $\pm$ 0.06  & 0.574 $\pm$ 0.08 \Tstrut \\
vRNN 1L ~\cite{Castrejn2019ImprovedCV} & 10 & \ \ 682.08           & 0.304 $\pm$ 0.10  & 0.609 $\pm$ 0.11 \\
Hier-vRNN ~\cite{Castrejn2019ImprovedCV} & 10 & \ \ 567.51           & 0.264 $\pm$ 0.07  & 0.628 $\pm$ 0.10 \\
GHVAE~\cite{wu2021greedy}              & 10 & \ \ 418.00 & 0.193 $\pm$ 0.014 & \textbf{0.740} $\pm$ 0.04 \\
\textbf{MCVD} spatin past-mask (Ours)                  &  \textbf{5} &              \ \ 184.81      &     0.121 $\pm$ 0.05              & 0.720 $\pm$ 0.11 \\ 
\textbf{MCVD} concat past-mask (Ours)                  &  \textbf{5} &              \ \ \textbf{141.31}      &     \textbf{0.112} $\pm$ 0.05              & 0.690 $\pm$ 0.12 \\ 
\bottomrule
\end{tabular}
\end{table}

%



\section{Conclusion}

We have shown how to obtain SOTA video prediction and interpolation results with randomly masked conditional video diffusion models using a relatively simple architecture. We found that past-masking was able to improve performance across all model variants and configurations tested. 
We believe our approach may pave the way forward toward high quality larger-scale video generation.
\begin{wraptable}[10]{r}{.6\textwidth}
\vspace{1em}
    \small
	\caption{Unconditional generation of BAIR video frames.}
	\label{tab:bair_gen}
	\centering
	\begin{tabular}{l|c|c}
	\toprule
		\textbf{BAIR} ($64\times64$) [0 $\rightarrow$ $pred$; trained on 5] & $pred$ & FVD$\downarrow$ \\
		\cmidrule(){1-3}
		\textbf{MCVD} spatin past-mask (Ours) & 16 & 267.8 \\
		\textbf{MCVD} concat past-mask (Ours) & 16 & \textbf{228.5} \\
		\cmidrule(){1-3}
		\textbf{MCVD} spatin past-mask (Ours) & 30 & 399.8 \\
		\textbf{MCVD} concat past-mask (Ours) & 30 & \textbf{348.2} \\
		\bottomrule
	\end{tabular}
\end{wraptable}
\begin{wraptable}[8]{r}{.6\textwidth}
\vspace{-1em}
    \small
	\centering
	\caption{Unconditional generation of UCF-101 video frames.}
	\begin{tabular}{l|r|c}
	\toprule
		\textbf{UCF-101} ($64\times64$)  [0 $\rightarrow$ 16; trained on $k$] & $k$ & FVD$\downarrow$ \\
		\cmidrule(){1-3}
		MoCoGAN-MDP \citep{yushchenko2019markov} & 16 & 1277.0 \\
		\textbf{MCVD} concat past-mask (Ours) & \textbf{4} & 1228.3  \\
		TGANv2 \citep{saito2020train} & 16 & 1209.0  \\
		\textbf{MCVD} spatin past-mask (Ours) & \textbf{4} & 1143.0 \\
		DIGAN \citep{yu2022generating} & 16 & \textbf{655.0} \\
		\bottomrule
	\end{tabular}
	\label{tab:ucf_gen}
\end{wraptable}
\textbf{Limitations.} Videos generated by these models are still small compared to real movies, and they can still become blurry or inconsistent when the number of generated frames is very large. Our unconditional generation results on the highly diverse UCF-101 dataset are still far from perfect. More work is clearly needed to scale these models to larger datasets with more diversity and with longer duration video. As has been the case in many other settings, simply using larger models with many more parameters is a strategy that is likely to improve the quality and flexibility of these models -- we were limited to 4 GPUs for our work here. There is also a need for faster sampling methods capable of maintaining quality over time. %
%
%

Given our strong interpolation results, conditional diffusion models which generate skipped frames could make it possible to generate much longer, but consistent video through a strategy of first generating sparse distant frames in a block, followed by an interpolative diffusion step for the missing frames. 
%

\begin{table}[!bth]
\small
\caption{Video Interpolation results (64 $\times$ 64). Given $p$ past + $f$ future frames $\rightarrow$ interpolate $k$ frames. Reporting average of the best metrics out of $n$ trajectories per test sample. $\downarrow\!(p\!+\!f)$ and $\uparrow\!k$ is harder. We used MCVD spatin past-mask for SMMNIST and KTH, and MCVD concat past-future-mask for BAIR. We also include results on SMMNIST for a "pure" model trained without any masking.}
\vspace{1em}
\label{tab:interp}
\setlength{\tabcolsep}{1pt}
\centering
\resizebox{\textwidth}{!}{
\begin{tabular}{l||ccc|cc||ccc|cc||ccc|cc}
\toprule
 & \multicolumn{5}{c||}{\textbf{SMMNIST} ($64\times64$)} & \multicolumn{5}{c||}{\textbf{KTH} ($64\times64$)} & \multicolumn{5}{c}{\textbf{BAIR} ($64\times64$)} \\
 & $p\!\!+\!\!f$ & $k$ & $n$ & PSNR$\uparrow$      & SSIM$\uparrow$
 & $p\!\!+\!\!f$ & $k$ & $n$ & PSNR$\uparrow$      & SSIM$\uparrow$
 & $p\!\!+\!\!f$ & $k$ & $n$ & PSNR$\uparrow$      & SSIM$\uparrow$ \\ \hline
SVG-LP~\scriptsize{\cite{denton2018SVGLP}} & 18 & 7 & 100 & 13.543 & 0.741 & 18 & 7 & 100 & 28.131 & 0.883 & 18 & 7 & 100 & 18.648 & 0.846  \Tstrutsmall  \\
FSTN~\footnotesize{\cite{lu2017flexible}} & 18 & 7 & 100 & 14.730 & 0.765 & 18 & 7 & 100 & 29.431 & 0.899 & 18 & 7 & 100 & 19.908 & 0.850 \\
SepConv~\footnotesize{\cite{niklaus2017video}} & 18 & 7 & 100 & 14.759 & 0.775 & 18 & 7 & 100 & 29.210 & 0.904 & 18 & 7 & 100 & 21.615 & 0.877    \\
SuperSloMo~\footnotesize{\cite{jiang2018super}} & 18 & 7 & 100 & 13.387 & 0.749 & 18 & 7 & 100 & 28.756 & 0.893 & -- & -- & -- & -- & -- \\
SDVI full~\footnotesize{\cite{xu2020stochastic}} & 18 & 7 & 100 & 16.025 & 0.842 & 18 & 7 & 100 & 29.190 & 0.901 & 18 & 7 & 100 & 21.432 & 0.880    \\
SDVI~\footnotesize{\cite{xu2020stochastic}} & 16 & 7 & 100 & 14.857 & 0.782 & 16 & 7 & 100 & 26.907 & 0.831 & 16 & 7 & 100 & 19.694 & 0.852    \\\hline
\multirow{3}{*}{\textbf{MCVD} (Ours)} & \textbf{10} & \textbf{10} & 100 & 20.944 & 0.854 & \textbf{15} & \textbf{10} & 100 & 34.669 & 0.943 & \textbf{4} & \textbf{5} & 100 & 25.162 & 0.932  \Tstrut  \\
& \textbf{10} & \textbf{5} & \textbf{10} & 27.693 & 0.941 & \textbf{15} & \textbf{10} & \textbf{10} & 34.068 & 0.942 & \textbf{4} & \textbf{5} & \textbf{10} & 23.408 & 0.914   \\
& \multicolumn{3}{c|}{pure} & 18.385 & 0.802 & \textbf{10} & \textbf{5} & \textbf{10} & 35.611 & 0.963 &  &  &  &  & \\
\bottomrule
\end{tabular}
}
\end{table}

\textbf{Broader Impacts.} 
High-quality video generation is potentially a powerful technology that could be used by malicious actors for applications such as creating fake video content. Our formulation focuses on capturing the distributions of real video sequences. High-quality video prediction could one day find use in applications such as autonomous vehicles, where the cost of errors could be high. Diffusion methods have shown great promise for covering the modes of real probability distributions. In this context, diffusion-based techniques for generative modelling may be a promising avenue for future research where the ability to capture modes properly is safety critical. Another potential point of impact is the amount of computational resources being spent for these applications involving the high fidelity and voluminous modality of video data. We emphasize the use of limited resources in achieving better or comparable results. Our submission provides evidence for more efficient computation involving fewer GPU hours spent in training time.

\begin{ack}
We thank Digital Research Alliance of Canada for the GPUs which were used in this work. Alexia, Vikram thank their wives and cat for their support. We thank CIFAR for support under the AI Chairs program, and NSERC for support under the Discovery grants program, application ID 5018358. 
\end{ack}

\bibliographystyle{unsrtnat}
\interlinepenalty=10000
\bibliography{refs}


\newpage
\section*{Checklist}
\begin{enumerate}
\item For all authors...
\begin{enumerate}
  \item Do the main claims made in the abstract and introduction accurately reflect the paper's contributions and scope?
    \answerYes{}
  \item Did you describe the limitations of your work?
    \answerYes{}
  \item Did you discuss any potential negative societal impacts of your work?
    \answerYes{}
  \item Have you read the ethics review guidelines and ensured that your paper conforms to them?
    \answerYes{}
\end{enumerate}
\item If you are including theoretical results...
\begin{enumerate}
  \item Did you state the full set of assumptions of all theoretical results?
    \answerNA{}
        \item Did you include complete proofs of all theoretical results?
    \answerNA{}
\end{enumerate}
\item If you ran experiments...
\begin{enumerate}
  \item Did you include the code, data, and instructions needed to reproduce the main experimental results (either in the supplemental material or as a URL)?
    \answerYes{}{We will release the code with the supplementary}
  \item Did you specify all the training details (e.g., data splits, hyperparameters, how they were chosen)?
    \answerYes{}{We will provide this in the supplementary material, there was insufficient space to provide this in the main paper.}
        \item Did you report error bars (e.g., with respect to the random seed after running experiments multiple times)?
    \answerYes{Whenever possible.}
        \item Did you include the total amount of compute and the type of resources used (e.g., type of GPUs, internal cluster, or cloud provider)?
    \answerYes{}
\end{enumerate}
\item If you are using existing assets (e.g., code, data, models) or curating/releasing new assets...
\begin{enumerate}
  \item If your work uses existing assets, did you cite the creators?
    \answerYes{}
  \item Did you mention the license of the assets?
    \answerNo{Readers can refer to the original references.}
  \item Did you include any new assets either in the supplemental material or as a URL?
    \answerNo{However we will link to a URL after peer review.}
  \item Did you discuss whether and how consent was obtained from people whose data you're using/curating?
    \answerNA{}
  \item Did you discuss whether the data you are using/curating contains personally identifiable information or offensive content?
    \answerNA{We are using publicly available data collected by others.}
\end{enumerate}
\item If you used crowdsourcing or conducted research with human subjects...
\begin{enumerate}
  \item Did you include the full text of instructions given to participants and screenshots, if applicable?
    \answerNA{}
  \item Did you describe any potential participant risks, with links to Institutional Review Board (IRB) approvals, if applicable?
    \answerNA{}
  \item Did you include the estimated hourly wage paid to participants and the total amount spent on participant compensation?
    \answerNA{}
\end{enumerate}
\end{enumerate}

\newpage
\appendix

\section{Appendix}

We provide some additional information regarding model size, memory requirements, batch size and computation times in \autoref{tab:memory}. This is followed by additional results and visualizations for SMMNIST, KTH, BAIR, UCF-101 and Cityscapes.

For more sample videos, please visit \url{https://mask-cond-video-diffusion.github.io}
For the code and pre-trained models, please visit \url{https://github.com/voletiv/mcvd-pytorch}. Our MCVD concat past-future-mask and past-mask results are of particular interest as they yield SOTA results across many benchmark configurations. 

We tried to add the older FID and IS metrics (as opposed to the newer FVD metric which we used above) for UCF-101 as proposed in \citet{saito2017temporal}, but we
had difficulties integrating the chainer \citep{tokui2019chainer} based implementation of these metrics into our PyTorch \citep{paszke2019pytorch} code base.


\subsection{Computational requirements}

\begin{table}[h!]
\small
\caption{Compute used. "steps" indicates the checkpoint with the best approximate FVD, "GPU hours" is the total training time up to "steps".}
\label{tab:memory}
\begin{tabular}{l|rrrrrrr}
\toprule
Dataset, & params & CPU mem & batch & GPU & GPU mem & steps & GPU \\ 
model & & (GB) & size & & (GB) & & hours \\ \hline
SMMNIST concat & 27.9M & 3.6 & 64 & Tesla V100 & 14.5 & 700000  & 78.9 \\
SMMNIST spatin & 53.9M & 3.3 & 64 & RTX 8000 & 23.4  & 140000  & 39.7 \\
KTH concat & 62.8M & 3.2 & 64 & Tesla V100 & 21.5  & 400000  & 65.7 \\
KTH spatin & 367.6M & 8.9 & 64 & A100 & 145.9  & 340000  & 45.8 \\
BAIR concat & 251.2M & 5.1 & 64 & Tesla V100 & 76.5  & 450000  & 78.2 \\
BAIR spatin & 328.6M & 9.2 & 64 & A100 & 86.1  & 390000  & 50.0 \\
Cityscapes concat & 262.1M & 6.2 & 64 & Tesla V100 & 78.2  & 900000  & 192.83 \\
Cityscapes spatin & 579.1M & 8.9 & 64 & A100 & 101.2  & 650000  & 96.0 \\
UCF concat & 565.0M & 8.9 & 64 & Tesla V100 &  100.1 & 900000  & 183.95 \\
UCF spatin & 739.4M & 8.9 & 64 & A100 & 115.2  & 550000  & 79.5 \\
\bottomrule
\end{tabular}
\end{table}

\subsection{Stochastic Moving MNIST}

\begin{figure}[!htb]
\includegraphics[width=\linewidth]{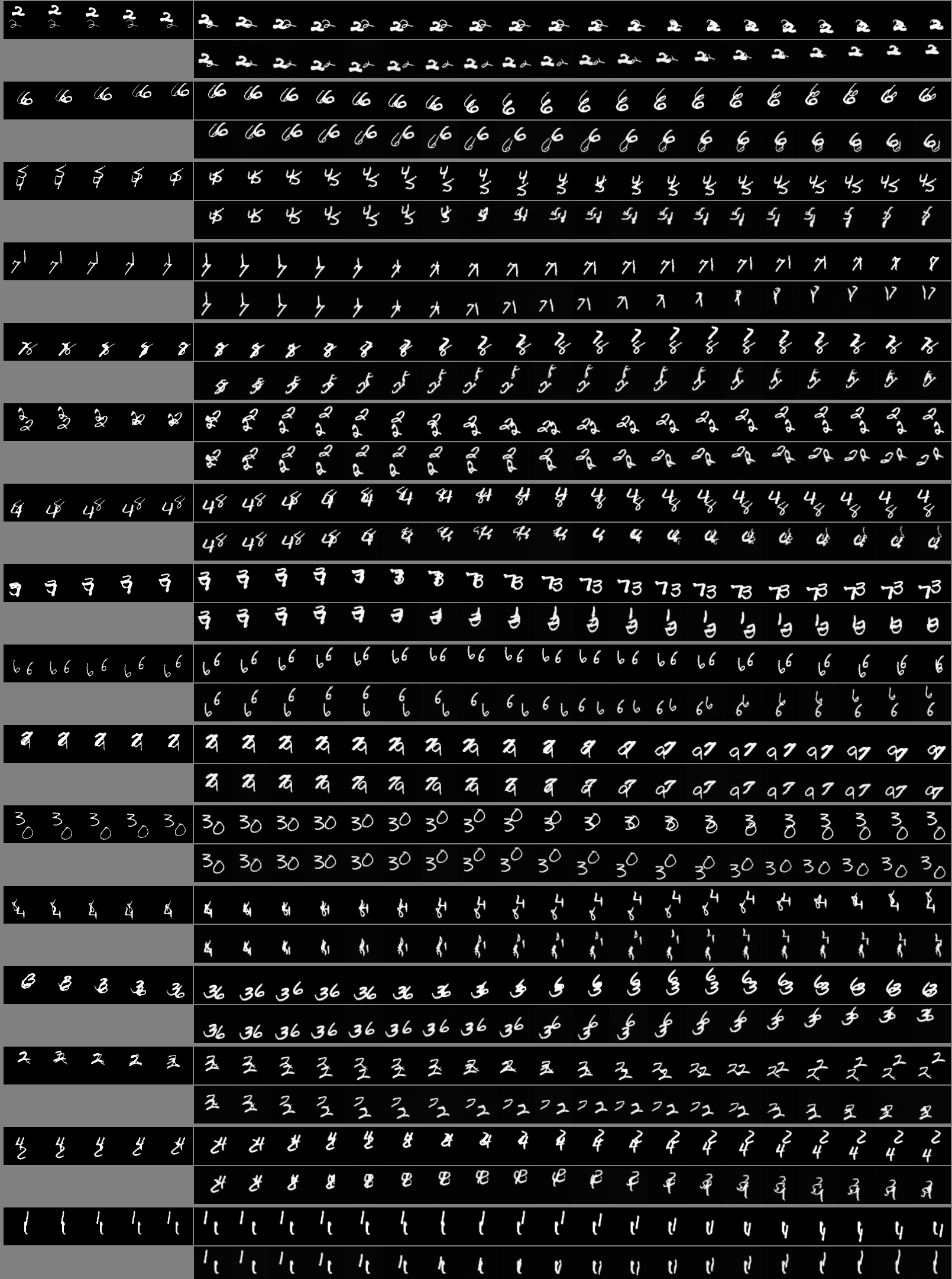}
\caption{SMMNIST 5 $\to$ 10, trained on 5 (prediction). For each sample, top row is real ground truth, bottom is predicted by our MCVD model.}
\label{fig:SMMNIST_2c5_images_1}
\end{figure}

In \autoref{tab:SMMNIST_ablation} we provide results for more configurations of our proposed approach on the SMMNIST evaluation. In \autoref{fig:SMMNIST_2c5_images_1} we provide some visual results for SMMNIST.

\begin{table}[h!]
\small
\caption{Results on the SMMNIST evaluation, conditioned on 5 past frames, predicting 10 frames using models trained to predict 5 frames at a time.}
\label{tab:SMMNIST_ablation}
\begin{tabular}{l|lllll}
\toprule
\textbf{SMMNIST} [5 $\rightarrow$ 10; trained on 5] & FVD$\downarrow$  & PSNR$\uparrow$  & SSIM$\uparrow$ & LPIPS$\downarrow$ & MSE$\downarrow$ \\ \hline
\textbf{MCVD} concat                  & 25.63 $\pm$ 0.69  &     17.22  &  0.786 & 0.117 & 0.024  \\
\textbf{MCVD} concat past-future-mask & \textbf{20.77} $\pm$ 0.77  &     16.33  & 0.753  & 0.139  &  0.028  \\
\textbf{MCVD} spatin                  & 23.86 $\pm$ 0.67  &     17.07  &  0.785 & 0.129 & 0.025  \\
\textbf{MCVD} spatin future-mask        & 44.14 $\pm$ 1.73  &     16.31  &  0.758 & 0.141 & 0.027    \\
\textbf{MCVD} spatin past-future-mask & 36.12 $\pm$ 0.63 & 16.15 & 0.748 & 0.146 & 0.027  \\
\bottomrule
\end{tabular}
\end{table}

\clearpage

\subsection{KTH}
\begin{table}[h!]
\small
\caption{Full table: Results on the KTH evaluation, predicting, 30 and 40 frames using models trained to predict $k$ frames at a time. All models condition on 10 past frames. SSIM numbers are updated from the main text after fixing a bug in the calculation. }
\label{tab:KTH_appendix}
\begin{tabular}{l|rc|lll}
\toprule
\textbf{KTH} [10 $\rightarrow$ $pred$; trained on $k$] & $k$ & $pred$ & FVD$\downarrow$    & PSNR$\uparrow$  & SSIM$\uparrow$  \\ \hline
SV2P~\citep{Babaeizadeh2017StochasticVV} & 10 & 30 & 636 $\pm$ 1  & 28.2  & 0.838               \\
SVG-LP~\citep{denton2018SVGLP}      & 10 & 30 & 377 $\pm$ 6 & 28.1 & 0.844 \\
SAVP~\citep{Lee2018StochasticAV}         & 10 & 30 & 374 $\pm$ 3 & 26.5  & 0.756 \\
\textbf{MCVD} spatin (Ours) & \textbf{5} & 30 & 323 $\pm$ 3 & 27.5   &  0.835    \\
\textbf{MCVD} concat past-future-mask (Ours) & \textbf{5} & 30 & 294.9 & 24.3   &  0.746    \\
SLAMP~\citep{akan2021slamp}              & 10 & 30 & 228 $\pm$ 5 & 29.4 & 0.865 \\
SRVP~\citep{franceschi2020stochastic}    & 10 & 30 & 222 $\pm$ 3 & 29.7 & 0.870 \\
\hline
Struct-vRNN~\citep{minderer2019unsupervised}            & 10 & 40   & 395.0  & 24.29 & 0.766 \\
\textbf{MCVD} concat past-future-mask (Ours) & \textbf{5} & 40 & 368.4 & 23.48   &  0.720    \\
\textbf{MCVD} spatin (Ours)                                    & \textbf{5} & 40   & 331.6 $\pm$ 5  &     26.40  &  0.744    \\
\textbf{MCVD} concat (Ours)                                    & \textbf{5} & 40   & 276.6 $\pm$ 3 &     26.20  &  0.793    \\
SV2P time-invariant~\citep{Babaeizadeh2017StochasticVV} & 10 & 40   & 253.5 & 25.70 & 0.772 \\
SV2P time-variant ~\citep{Babaeizadeh2017StochasticVV}  & 10 & 40   & 209.5 & 25.87 & 0.782 \\
SAVP~\citep{Lee2018StochasticAV}                        & 10 & 40   & 183.7 & 23.79 & 0.699 \\
SVG-LP~\citep{denton2018SVGLP}                     & 10 & 40   & 157.9 & 23.91 & 0.800 \\
SAVP-VAE~\citep{Lee2018StochasticAV}                    & 10 & 40   & 145.7 & 26.00 & 0.806 \\
Grid-keypoints~\citep{gao2021accurate}                  & 10 & 40   & 144.2 & 27.11 & 0.837 \\
\bottomrule
\end{tabular}
\end{table}

\clearpage

\begin{figure}[htb!]
\includegraphics[width=\linewidth]{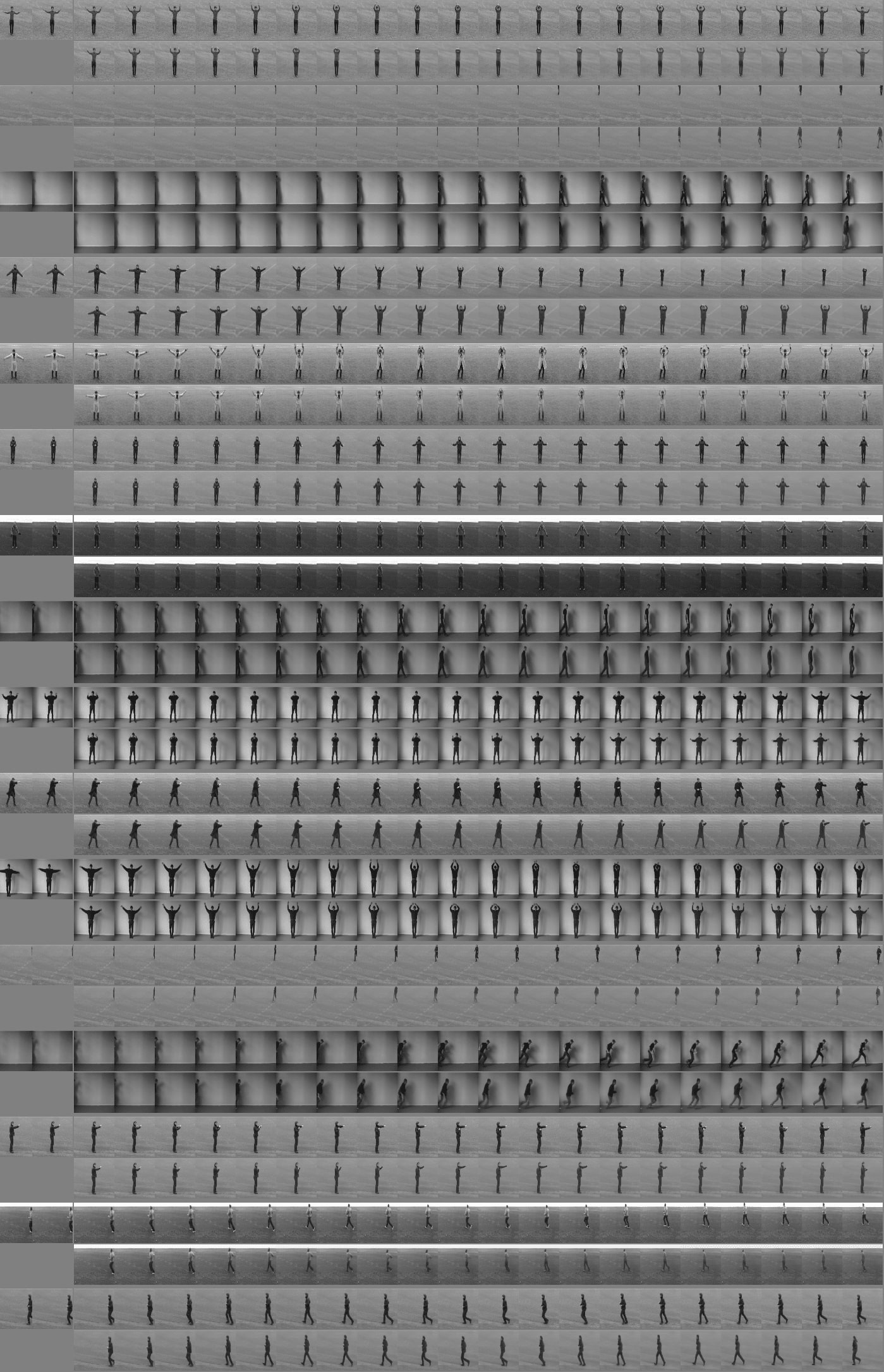}
\caption{\textbf{KTH} 5 $\to$ 20, trained on 5 (prediction). For each sample, top row is real ground truth, bottom is predicted by our MCVD model. (We show only 2 conditional frames here)}
\label{fig:KTH_5c10_images}
\end{figure}


\clearpage

\subsection{BAIR}

\begin{threeparttable}[h!]
    \small
	\caption{Full table: Results on the BAIR evaluation conditioning on $p$ past frames and predict pr frames in the future, using models trained to predict $k$ frames at at time. 
	}
	\label{tab:bair_pred_appendix}
	\centering
	\begin{tabular}{l|crc|rcc}
	\toprule
		BAIR [past (p) $\rightarrow$ pred (pr) ; trained on k] & p & k & pr & FVD$\downarrow$ & PSNR$\uparrow$ & SSIM$\uparrow$ \\
		\cmidrule(){1-7}
        LVT \citep{rakhimov2020latent} & 1 & 15 & 15 & 125.8 & -- & -- \\
        DVD-GAN-FP \citep{clark2019adversarial} & 1 & 15 & 15 & 109.8 & -- & -- \\
		\textbf{MCVD} spatin (Ours) & 1 & \textbf{5} & 15 & 103.8 & 18.8 & 0.826  \\
        TrIVD-GAN-FP \citep{luc2020transformation} & 1 & 15 & 15 & 103.3 & -- & -- \\
		VideoGPT \citep{yan2021videogpt} & 1 & 15 & 15 & 103.3 & -- & -- \\
		CCVS \citep{le2021ccvs} & 1 & 15 & 15 & 99.0 & -- & -- \\
	    \textbf{MCVD} concat (Ours) & 1 & \textbf{5} & 15  & 98.8 & 18.8 & 0.829 \\
		\textbf{MCVD} spatin past-mask (Ours) & 1 & \textbf{5} & 15 & 96.5 & 18.8 & 0.828 \\
		\textbf{MCVD} concat past-mask (Ours) & 1 & \textbf{5} & 15 & 95.6 & 18.8 & 0.832  \\
        Video Transformer \citep{weissenborn2019scaling} & 1 & 15 & 15 & 94-96\tnote{a} & -- & -- \\
		FitVid \citep{babaeizadeh2021fitvid} & 1 & 15 & 15 & 93.6 & -- & -- \\
		\textbf{MCVD} concat past-future-mask (Ours) & 1 & \textbf{5} & 15 & \textbf{89.5} & 16.9 & 0.780  \\
		\cmidrule(){1-7}
        SAVP \citep{Lee2018StochasticAV} & 2 & 14 & 14 & 116.4 & -- & -- \\
		\textbf{MCVD} spatin (Ours) & 2 & \textbf{5} & 14 & 94.1 & 19.1 & 0.836  \\
		\textbf{MCVD} spatin past-mask (Ours) & 2 & \textbf{5} & 14 & 90.5 & 19.2 & 0.837  \\
		\textbf{MCVD} concat (Ours)  & 2 & \textbf{5} & 14 & 90.5 & 19.1 & 0.834  \\
		\textbf{MCVD} concat past-future-mask (Ours) & 2 & \textbf{5} & 14 & 89.6 & 17.1 & 0.787  \\
		\textbf{MCVD} concat past-mask (Ours) & 2 & \textbf{5} & 14 & \textbf{87.9} & 19.1 & 0.838  \\
		\cmidrule(){1-7}
		SVG-LP~\citep{akan2021slamp}    & 2    & 10   & 28   & 256.6          &      --            & 0.816    \\
        SVG~\citep{akan2021slamp}.   & 2 & 12 & 28 & 255.0 & 18.95 & 0.8058 \\
        SLAMP \citep{akan2021slamp} & 2 & 10 & 28 & 245.0 & 19.7 & 0.818  \\
        SRVP \citep{franceschi2020stochastic} & 2 & 12 & 28 & 162.0 & 19.6 & 0.820 \\
		WAM \citep{jin2020exploring} & 2 & 14 & 28 & 159.6 & 21.0 & 0.844 \\
        SAVP~\citep{Lee2018StochasticAV} & 2 & 12 & 28 & 152.0 & 18.44 & 0.7887 \\
        vRNN 1L~\cite{Castrejn2019ImprovedCV}                                & 2    & 10   & 28   & 149.2          &        --          & 0.829   \\
        SAVP~\citep{Lee2018StochasticAV} & 2    & 10   & 28   & 143.4          &        --          & 0.795  \\
        Hier-vRNN ~\citep{Castrejn2019ImprovedCV}                             & 2    & 10   & 28   & 143.4           &         --         & 0.822   \\
		
		\textbf{MCVD} spatin (Ours) & 2 & \textbf{5} & 28 & 132.1 & 17.5 & 0.779  \\
		\textbf{MCVD} spatin past-mask (Ours) & 2 & \textbf{5} & 28 & 127.9 & 17.7 & 0.789 \\
		\textbf{MCVD} concat (Ours)  & 2 & \textbf{5} & 28 & 120.6 & 17.6 & 0.785  \\
		\textbf{MCVD} concat past-mask (Ours) & 2 & \textbf{5} & 28 & 119.0 & 17.7 & 0.797 \\
		\textbf{MCVD} concat past-future-mask (Ours) & 2 & \textbf{5} & 28 & \textbf{118.4} & 16.2 & 0.745 \\
		\bottomrule
	\end{tabular}
    \begin{tablenotes}
    \item[a] 94 on only the first frames, 96 on all subsquences of test frames
    \end{tablenotes}
\end{threeparttable}

\clearpage

\begin{figure}[htb!]
\includegraphics[width=\linewidth]{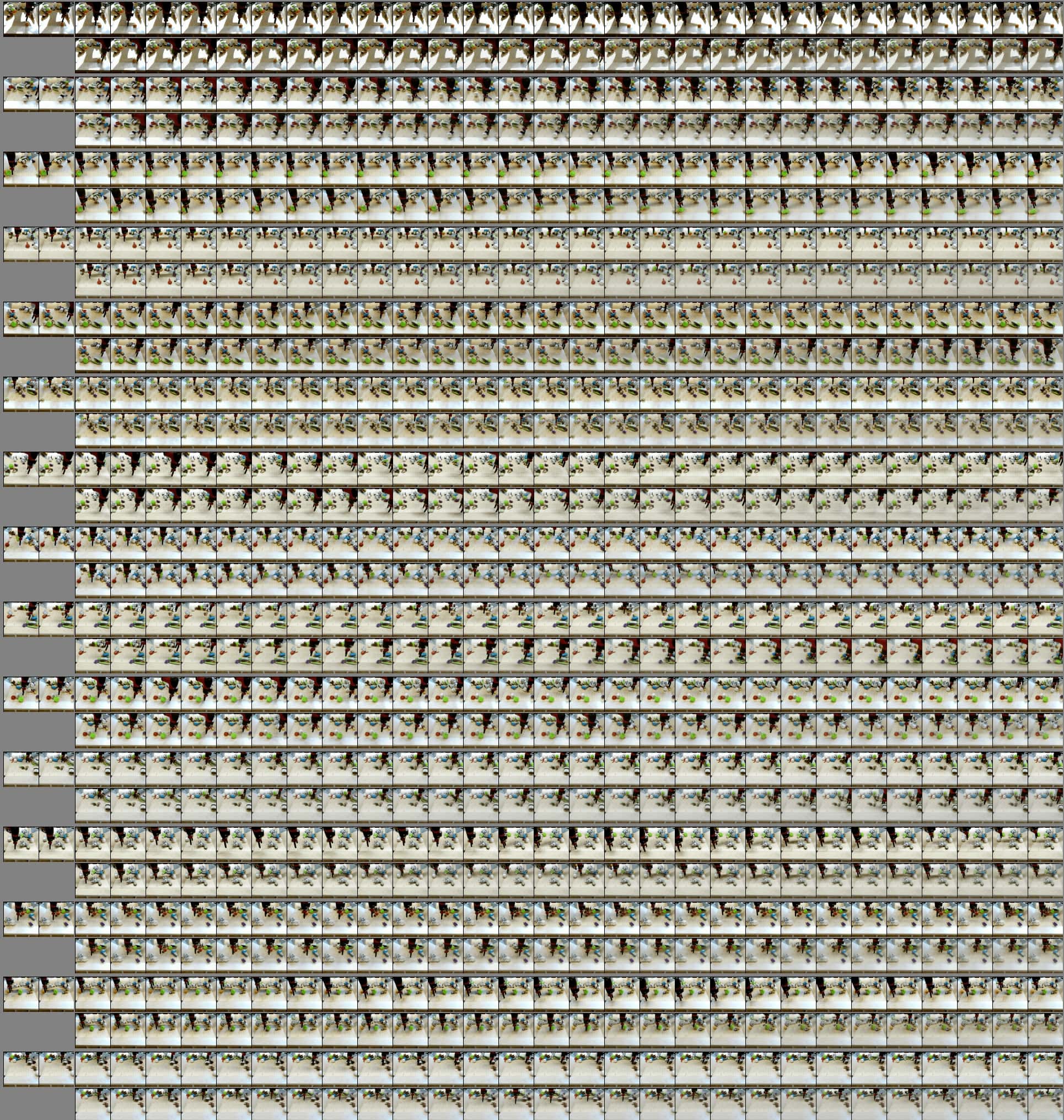}
\caption{\textbf{BAIR} 2 $\to$ 28, trained on 5 (prediction). For each sample, top row is real ground truth, bottom is predicted by our MCVD model.}
\label{fig:BAIR_2c2_images}
\end{figure}


\clearpage

\subsection{UCF-101}

\begin{figure}[htb!]
\includegraphics[width=\linewidth]{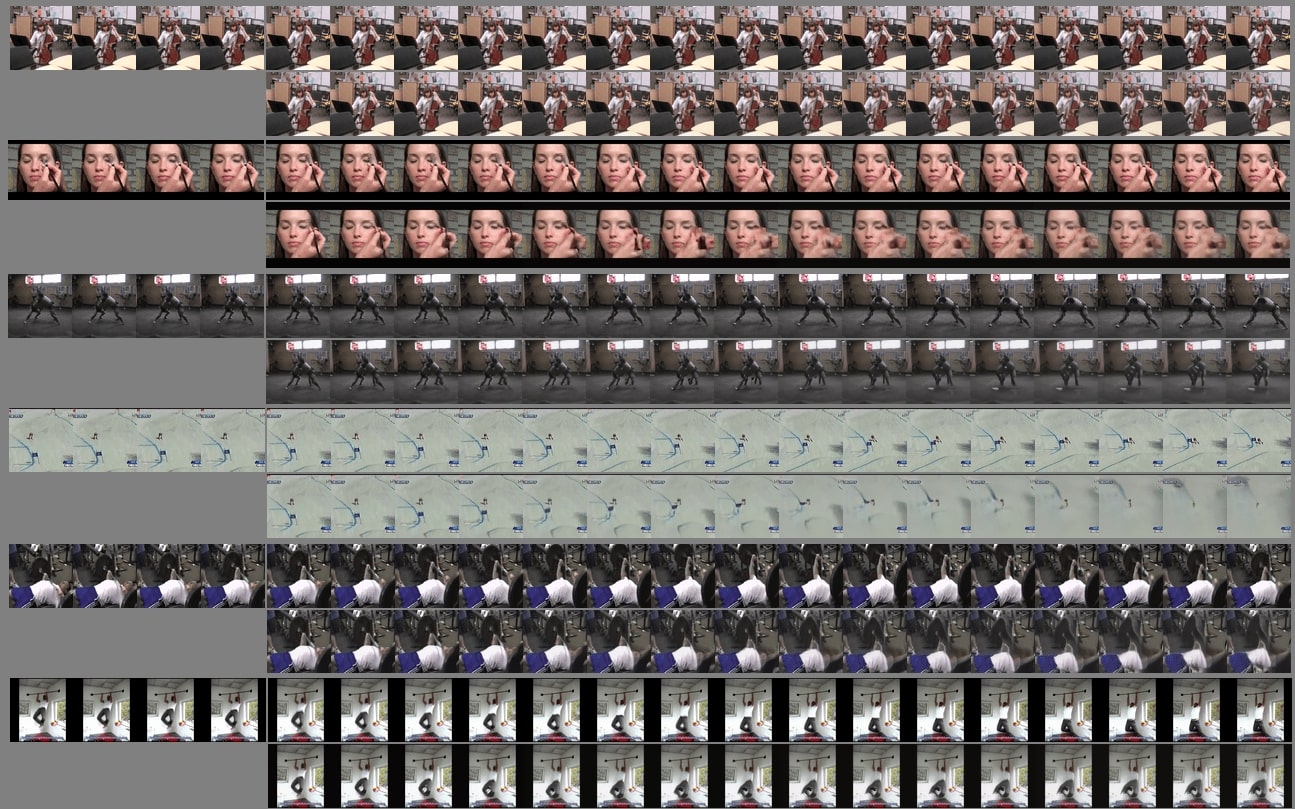}
\caption{\textbf{UCF-101} 4 $\to$ 16, trained on 4 (\textbf{prediction}). For each sample, top row is real ground truth, bottom is predicted by our MCVD model.}
\label{fig:UCF101_4c4_images_pred}
\end{figure}

\begin{figure}[htb!]
\includegraphics[width=\linewidth]{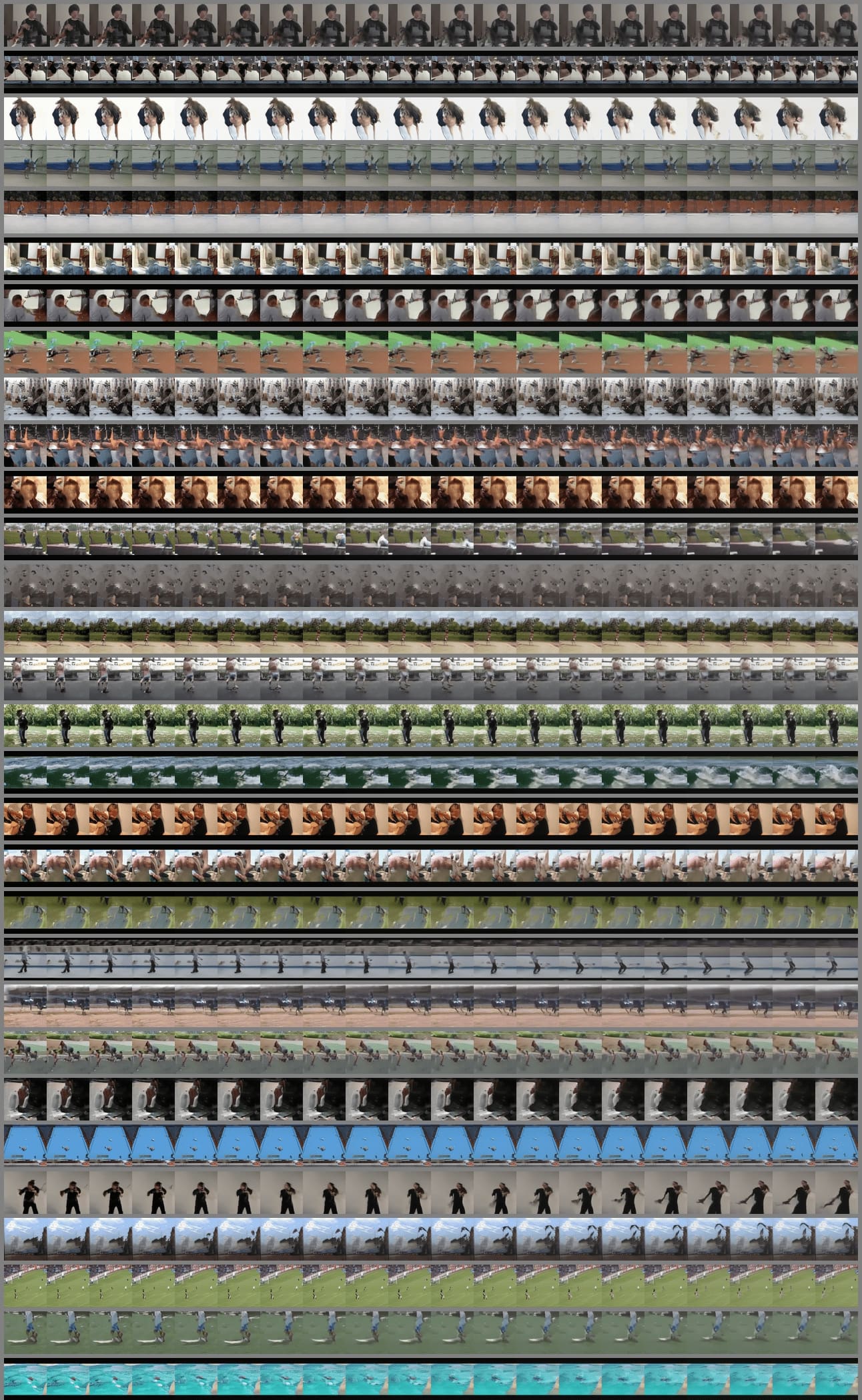}
\caption{\textbf{UCF-101}  0 $\to$ 4 (\textbf{generation})}
\label{fig:UCF101_4c4_images_gen}
\end{figure}

\clearpage

\subsection{Cityscapes}

Here we provide some examples of future frame prediction for Cityscapes sequences conditioning on two frames and predicting the next 7 frames.

\begin{figure}[h!]
    \centering
    \includegraphics[width=\textwidth]{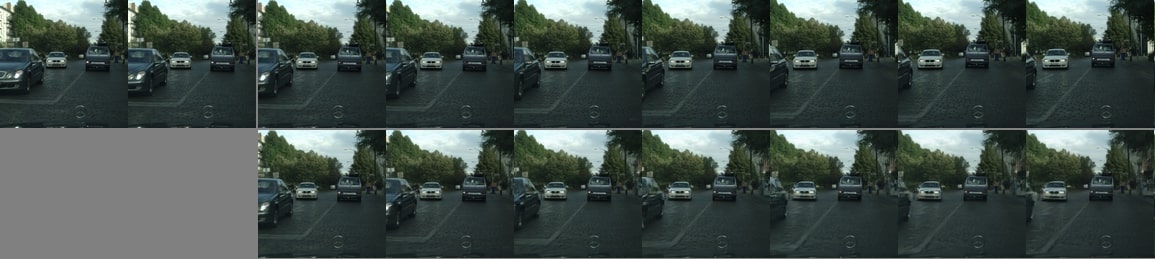}
    \includegraphics[width=\textwidth]{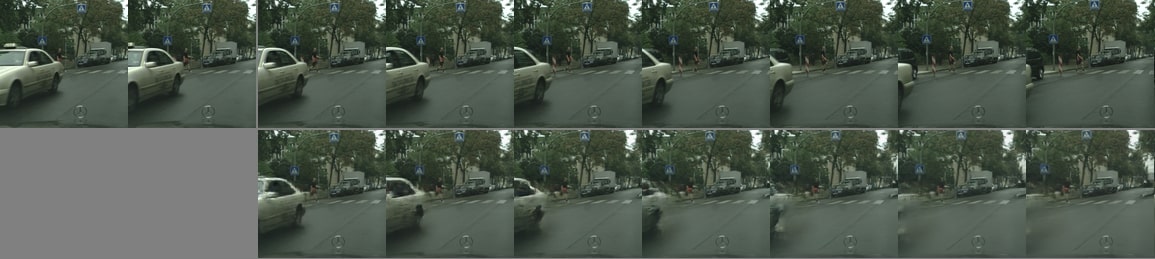}
    \includegraphics[width=\textwidth]{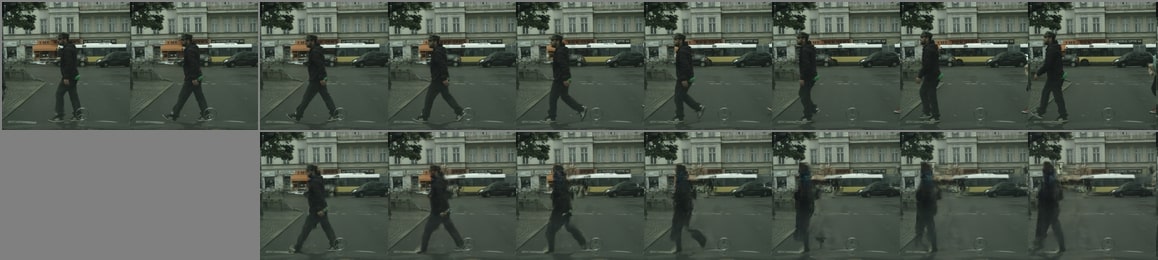}
    \includegraphics[width=\textwidth]{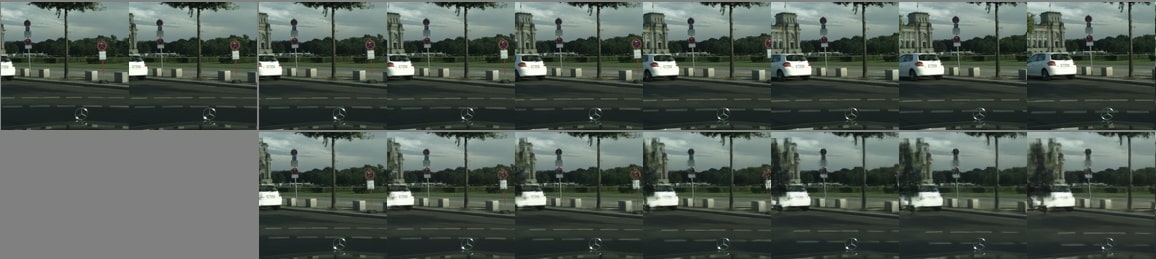}
    \caption{\textbf{Cityscapes}: 2 $\to$ 7, trained on 5 (prediction); Conditioning on the two frames in the top left corner of each block of two rows of images, we generate the next 7 frames. The top row is the true frames, bottom row contains the generated frames. We use the \textbf{MCVD} concat model variant.}
    \label{fig:more_city}
\end{figure}

For more examples, please visit \url{https://mask-cond-video-diffusion.github.io}

\end{document}